\documentclass[journal]{IEEEtran}
\usepackage{graphicx}
\usepackage{amsmath,amssymb}

\usepackage{multirow}

\usepackage{hyperref}
\usepackage{breakurl}

\ifCLASSINFOpdf
\else
\fi
\usepackage{algorithmic}
\begin{document}
%\title{Evolving Spectrum-diverse Networks with a Unified Neuroevolution Architecture}
\title{Spectrum-Diverse Neuroevolution with Unified Neural Models}
%\title{Evolving a Unified Neural Model with Spectrum-diverse Networks}

%\author{Michael~Shell,~\IEEEmembership{Member,~IEEE,}
%        John~Doe,~\IEEEmembership{Fellow,~OSA,}
%        and~Jane~Doe,~\IEEEmembership{Life~Fellow,~IEEE}% <-this % stops a space
%\thanks{M. Shell is with the Department
%of Electrical and Computer Engineering, Georgia Institute of Technology, Atlanta,
%\thanks{Manuscript received April 19, 2005; revised September 17, 2014.}}

\author{Danilo Vasconcellos Vargas and Junichi Murata
\thanks{D. V. Vargas is with the Graduate School of Information Science and Electrical Engineering, Kyushu University, Fukuoka, Japan (email: vargas@cig.ees.kyushu-u.ac.jp)}
\thanks{J. Murata is with the Faculty of Information Science and Electrical Engineering, Kyushu University, Fukuoka, Japan (email: murata@cig.ees.kyushu-u.ac.jp)}
}


%\markboth{Journal of \LaTeX\ Class Files,~Vol.~13, No.~9, September~201}%
%{Shell \MakeLowercase{\textit{et al.}}: Bare Demo of IEEEtran.cls for Journals}

%\markboth{Journal of \LaTeX\ Class Files,~Vol.~13, No.~9, September~2015}%
\markboth{IEEE TRANSACTIONS ON NEURAL NETWORKS AND LEARNING SYSTEMS,~Vol.~x, No.~x, September~20xx}%
{Vargas and Murata: Evolving Spectrum-diverse Networks with a Unified Neuroevolution Architecture}



\maketitle


%lgorithm to face many classes of problems automatically, a more powerful representation is needed.
\begin{abstract}
Learning algorithms are being increasingly adopted in various applications.
However, further expansion will require methods that work more automatically.
To enable this level of automation, a more powerful solution representation is needed.
However, by increasing the representation complexity a second problem arises. The search space becomes huge and therefore an associated scalable and efficient searching algorithm is also required.
To solve both problems, first a powerful representation is proposed that unifies most of the neural networks features from the literature into one representation.
Secondly, a new diversity preserving method called Spectrum Diversity is created based on the new concept of chromosome spectrum that creates a spectrum out of the characteristics and frequency of alleles in a chromosome.
The combination of Spectrum Diversity with a unified neuron representation enables the algorithm to either surpass or equal NeuroEvolution of Augmenting Topologies (NEAT) on all of the five classes of problems tested.
Ablation tests justifies the good results, showing the importance of added new features in the unified neuron representation.
Part of the success is attributed to the novelty-focused evolution and good scalability with chromosome size provided by Spectrum Diversity.
Thus, this study sheds light on a new representation and diversity preserving mechanism that should impact algorithms and applications to come.\\

To download the code please access the following \href{https://github.com/zweifel/Physis-Shard}{link}.
\end{abstract}
%Here a topology and weight evolving artificial neural network is proposed to take full advantage of the unified neural network representation and Spectrum Diversity.
%Spectrum diversity is more scalable than speciation because of its use of chromosome spectra instead of the genes themselves.
%Moreover, novel enough chromosomes are kept even when their fitness is abismally poor, this allows different approaches to solve the problem to develop without the deleterious competition of faster evolving ones. 


% \fi
\IEEEpeerreviewmaketitle

\section{Introduction}

Learning algorithms are the only solution available to problems that are too complex to be solved with hand-coded programs.
They are, for example, used to recognize speech in cell phones, aid the trade in stock markets, process and extract knowledge from problems with big data, play games comparatively to humans and so on.
Moreover, they are a natural solution to problems that are expected to change over time, since they can learn, i.e. adapt, to changes.
%This kind of innate capability is still not used but would lead to a new generation of algorithms.

The impact of learning algorithms nowadays is enormous, yet the area has the potential to expand much further.
For that, however, it is necessary that learning algorithms be able to work on new problems without the presence of an expert.
Current algorithms behave sometimes erratically on new problems or require a lot of trial and error as well as prior knowledge to work.

This motivates us to create algorithms that can learn under problems never seen before and without any prior knowledge.
Moreover, this type of investigation have not only a socio-economic value but also a scientific one.
Learning can be defined as the ability of improving computational models (representations) to solve problems.
It is the idea of starting from zero or little prior knowledge and arriving at a reasonably good solution.
Naturally, the more prior knowledge is inserted into a learning algorithm, the less its learning capabilities are important.
%A learning algorithm with a lot of prior knowledge such as problem specific operators, modified parameters, problem specific structures and so on is not different from non-learning algorithms, losing many of the interesting properties of learning algorithms.
There is still much to be understood about the learning limitations and inherent trade-offs associated.
%There are also many open questions such as if t

To build an algorithm that can face very different problems it is necessary to have a general representation.
The most general representation is perhaps a graph with nodes composed of various types of functions.
Provided that brains are also graphs of neurons, this justifies the bio-inspiration in neural networks. 
The only way to optimize not only connection weights but also topology of neural networks is with neuroevolution, because evolution can optimize any kind of problem including ones where the model itself grows or decreases in size. 
Consequently, neuroevolution is chosen as the underlining basis for this work.

In this article, a neuroevolution based learning algorithm capable of learning various classes of problems without any prior-knowledge is proposed.
Moreover, this new neuroevolution based learning algorithm tries to unify most of the neuron representations into one.
Since every kind of neuron is specifically good for a certain type of task, the natural selection of the neuron type for each problem suits well the objective of solving a wide range of problems without any prior knowledge.
%This suits well the objective of solving a wide series of problems (mountain car, double pole balancing, non-Markov double pole balancing, multiplexer and function approximation) without any prior knowledge.
Furthermore, a new niching mechanism and diversity paradigm are created to aid the evolution of this complex representation. 
The proposed method is called Spectrum-diverse Unified Neuroevolution Architecture (SUNA).
In summary, SUNA\footnote{Download SUNA's Code: https://github.com/zweifel/Physis-Shard} has the following features:
\begin{itemize}
	\item Unified Neuron Representation - A new neuron representation is proposed that unifies most of the proposed neuron variations into one, giving the method a greater power of representation.
	\item Novelty-Based Diversity Preserving Mechanism - Solutions are not only different, but the frequency and types of neurons and connections used may suggest a completely different approach to solve the problem. For example, the use of many neuromodulated connections against normal connections in a network is a different approach. Such diversity in the approaches used should be preserved and therefore a new novelty-based diversity is proposed which keeps track of previous solutions and preserves the different approaches spotted. 
	\item Meaningful Diversity for Large Chromosomes- Instead of comparing chromosomes, this diversity measure compares values of a set of phenotype-based or genotype-based metrics. In other words, spectra\footnote{The word spectrum here is used to denote a set of values of histograms or metrics, defining the overall characteristics of the chromosome in a compressed manner.} of features are compared. By comparing spectra, the diversity is still meaningful for large chromosomes\footnote{High dimensional spaces are exponentially more sparse than low dimensional ones. That is the reason why distance metrics for long chromosomes are less meaningful. To avoid this complication, here the dimensions are converted to a small number of features which compose the spectrum of the chromosome} as well as it is problem independent.
	\item State-of-the-Art Results - In all five classes of problems tested, the proposed algorithm had either better or similar results when compared with NEAT.
	\item Few Parameters - SUNA has only eight parameters against $33$ of the NEAT algorithm.
\end{itemize}

\section{Neuroevolution}

Neuroevolution is an area of research resulting from the combination of the representation power of artificial neural networks \cite{hornik1989multilayer, haykin2004comprehensive} with the optimization capabilities of evolutionary methods.
For example, instead of using the usual gradient descent methods like backpropagation an evolutionary algorithm is used to train the neural network.

Many works investigated the evolution of fixed-topology neural networks \cite{whitley1993genetic, wieland1991evolving, saravanan1995evolving, moriarty1997forming}.
However, it is hard to specify a good representation in advance.
First, the complexity of the representation should match the complexity of the problem, therefore a completely automated algorithm is impossible.
Second, although it is possible to represent any function with a single hidden layer, it does not mean that any function is equally easy to represent with a single hidden layer. 
With the increase of hidden neurons, the search space becomes increasingly huge.
That is, the optimization algorithm has to deal not only with plateaus and difficult fitness landscapes, but also with an additional difficulty in the representation because many parameters are present as well as competing conventions (different neural networks that compute the same and exact function).

Deep Neural Networks (DNN) solves the second mentioned problem, i.e., by using many layers it is possible to learn representations in higher levels of abstractions \cite{schmidhuber2015deep},\cite{lecun2015deep}.
This justifies why this type of network sets records in image recognition \cite{krizhevsky2012imagenet},\cite{szegedy2014going} and speech recognition \cite{mikolov2011strategies},\cite{hinton2012deep} as well as has good results in many other types of applications.
Still, in order for DNNs to work specialists are need to set the network parameters and design its topology.
 
Recently, motivated by the limitations of fixed-topology approaches, there is an increasing interest in algorithms that evolve both the topology as well as the weights.
To these algorithms it is given the name of TWEANN (Topology and Weight Evolving Artificial Neural Network).
Some of the most famous direct encoding TWEANNs are Cellular Encoding \cite{gruau1994neural}, GNARL \cite{angeline1994evolutionary}, NEAT \cite{stanley2002evolving}, EANT \cite{kassahun2005efficient} and EPNet \cite{yao1997new}.
Naturally, some of them have also indirect encoding versions while others can be extended to indirect encoding without problems.
For a detailed review of TWEANNs please refer to \cite{floreano2008neuroevolution}.

%It is hypothesized that RBF nodes which local activation functions could let NEAT have a and then be able to solve fractured problems.

One of the most successful direct encoding TWEANNs is the NEAT algorithm which has many notable features that made it so popular.
We will explain these notable features as well as briefly describe how NEAT works in the following section.

\section{NEAT}

%The success of NEAT is due to three main ideas:
%\begin{itemize}
%	\item Historical markings that enable structure aware crossover;
%	\item 
%\end{itemize}

One of the central ideas of NEAT is the innovation number which works as a gene identification and tracking number.
Innovation number is an additional value attached to a gene.
%This value is generated every time a gene is created and is unique to each gene.
It enables the recognition of the same gene in different individuals, therefore it is possible to match the same genes while doing crossover.
Moreover, it is possible to avoid the problems associated with competing conventions during crossover.

NEAT uses a variable-size genetic encoding where each individual has a list of node genes and a list of connection genes:
\begin{itemize}
	\item Node genes specify if each of their associated nodes is an input, output or hidden node as well as state its innovation number;
	\item Connection genes determine connections between two nodes by specifying the in-node, out-node and weight of the connection. Moreover, there is an associated enable bit (whether or not this gene is expressed) and an innovation number.
\end{itemize}

\begin{figure}[!ht]
\centering
 \includegraphics[width=0.4\textwidth]{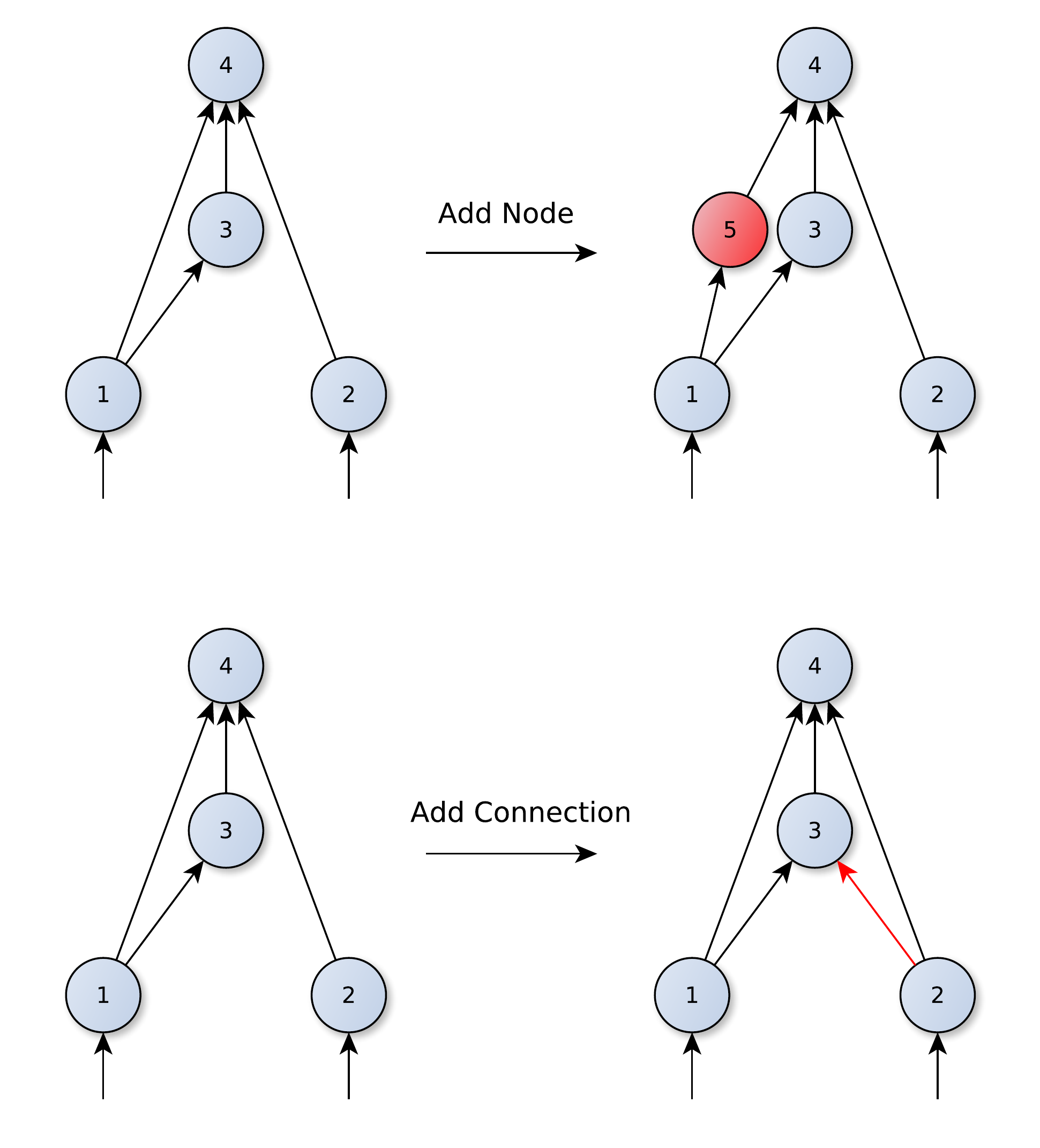}
 \caption{Structural mutations in NEAT.}
 \label{mutation_neat}
\end{figure}

There are four types of genetic operators in NEAT: two types of structural mutations, a simple weight mutation and crossover.
Structural mutations add complexity to the network by either including a new node in the middle of an existing connection or by adding a new connection between existing nodes (see Figure~\ref{mutation_neat}).
%Notice that the addition of the node in the middle of an existing connection prevents bloating.
Additionally, there is the usual weight mutation present in many other methods, where some individuals have the weight value of their connections either perturbed or set to a new random value.
Crossover is done by lining up the innovation numbers from both parents in chronological order.
Matching genes are inherited while unmatching ones are either inherited randomly when parents are equally fit or inherited from the fittest parent when one parent is fitter than the other.
This type of crossover avoids an expensive structural analysis of similarity while still preserving some similar structure present.
%(called disjoint when they are in the middle of othe matching genes or excess genes)

Different structures take different time to improve through evolution, e.g., complex structures take longer to improve than simpler ones.
Therefore, protecting different structures from unfair competition is necessary if one wants to have an unbiased evolution.
NEAT solves this by introducing speciation.
Each generation, individuals are separated into species by calculating a distance based on their matching and unmatching genes (i.e., same and different innovation numbers).
Individuals that have their distance below a given threshold are from the same species.
An explicit fitness sharing \cite{goldberg1987genetic} is used to make it harder for any species to grow and dominate the entire population.
Thus, species have a determined number of offspring slots proportional to the average fitness of its individuals while individuals compete with other individuals within the same species for the allocated offspring slots.
The entire population is replaced by its offspring.

NEAT start the search from a relatively small network where all inputs are connected to all outputs without hidden nodes.
Notice that the initial population diversity resides in the connection weights which are randomized rather than in the structure.
This biases the evolution towards minimal solutions while the genetic operators complexify the structure incrementally by adding more nodes and connections throughout the evolution process.

\section{Neural Networks' Features: A Unifying View}
%\section{Neural Networks' Features: A Unifying View Behind a Unified Neural Network Model}

%The area of artificial neural networks (ANNs) can be said to have began many years ago, when the first perceptron came to life.
%Motivations behind it were to solve reinforcement learning problems, but at that time end up being mostly used in classification problems.
%Nowadays ANNs are used i
%From an optimization perspective, artificial neural networks are computational models which have their parameters optimized by some optimization algorithm.
%Recently, not just the parameters are optimized but also the networks's structure, i.e. the network can grow (create neurons), shrink (delete neurons) and change connections (create or delete connections).

%\subsection{Features of Neural Networks}

%Neural networks have been researched for quite some time in many branches of the literature.
%In fact, given the fact that learning can be done in many ways and the models themselves can vary, it is difficult to have a broad picture of its many variations.
%This section wishes to mention most of the features present in current neural network models.
%Consequently, 
The number of types of neural networks is extremely vast.
Although they are called neural networks, they differ in so many aspects such as learning system, architecture\footnote{Architecture is defined here as features or dynamics of blocks of neurons such as bottlenecks in the network, connection probabilities, topology dynamics, etc.} and model that it is difficult to see them as a unity.
%They can differ in objective, for example self organizing map () to supervised ones such as multilayer perceptrons.
%We define here architecture as rules applied over blocks of neurons. 
Moreover, the features of the neural network used usually reflects the problem it is facing, therefore a unified neural network model is also required to automatically find the solution, independently of problem type.

%(for example, recurrent neural networks are used when the previous information is important).
%Therefore, to apply neural networks a knowledge of such features and their relation with the problems they can solve are needed.
%Things get more complicated when various systems need to be combined or parameters tweaked.

%The learning systems used are gradient descent based ones and other more robust optimization algorithms such as evolutionary algorithms.
%The architecture may differ in various ways, for example, bottlenecks in the network, connection probabilities, topology dynamics among others.
Motivated by the unification benefits, this section will focus on describing many model features in a unified way.
%, since giving the advantages non-fixed topology networks, the learning system choices should be constrained to evolutionary algorithms.
%Moreover, architecture's features are generic enough to be applied to any model, therefore .
%Let us classify the neural networks features into structural and logical ones.
Setting activation functions aside, allow us to divide the model features of a neural network into four broad aspects:
\begin{itemize}
	\item Different neuronal time scales;
	\item Inhibition and excitation of neurons;
	\item Synaptic plasticity;
	\item Feedback;
\end{itemize}

\subsection{Different Neuronal Time Scales}

Most neural networks use only one fixed fast time constant, which makes all neurons vary its output as fast as possible in relation to their input.
This is important feature in executing tasks which need fast reaction time.
However, a slower neuron can take into account a series of past inputs enabling it to decide over long-term events and even store information for other neurons.
The importance of time scales have been shown in many articles, for example, in self-localization of a neuroevolved robot in a corridor (the sensory-motor patterns in different time scales are used) \cite{nolfi2002evolving}, in reinforcement learning with multi-models and sub-goals \cite{precup1998multi}, with the Long Short-Term Memory (LSTM) Recurrent Neural Network (RNN) which use memory cells capable of storing/accumulating signals for a very long time (much more than usual RNN) until a forget signal is received \cite{gers2003learning}.
The different neuronal time scales even emerge naturally when a bottleneck is introduced in the neuron topology, creating a layer with long-term task related actions and another layer with short-term reactionary actions \cite{paine2005hierarchical}.

In fact, biological neurons have a membrane time constant that vary widely from neuron to neuron (the time constant depends on the type, density and regulation of the present ion channels) \cite{levitan2002neuron}.
Therefore, different time scales are present in biological neurons and experiments even show how this temporal hierarchy of neurons relates with the anatomical hierarchy \cite{kiebel2008hierarchy}.
In other words, the temporal hierarchy is an essential part of the brain.

\subsection{Inhibition and Excitation of Neurons}
\label{control_motivation}

\begin{figure}[ht]
\centering
 \includegraphics[width=0.4\textwidth]{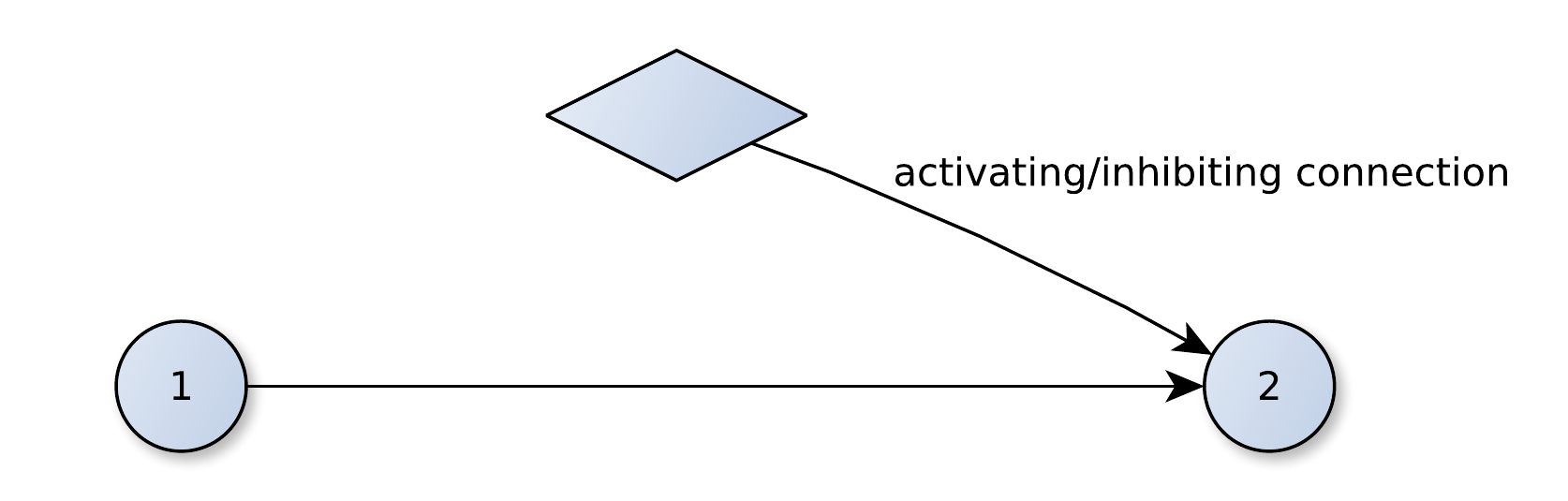}
 \caption{Neuromodulation of neurons is the activation/inhibition of neurons by another neuron. In the figure, the neuromodulatory neuron (represented by a diamond) is activating or inhibiting neuron $2$.}
 \label{nm1}
\end{figure}

Inhibition and excitation of neurons (the term ``neuromodulation of neurons" will also be used) is the ability of some neurons to inhibit (deactivate) or excite (activate) other neurons (see Figure~\ref{nm1}).
It is widely known that biological neural networks are heavily based on inhibition/excitation principles with groups of neurons influencing how other neurons will behave and if they are going or not going to activate \cite{kandel2012principles}, \cite{marder2002cellular}
\footnote{In biology, the inhibition/excitation of neurons by other neurons is also called neuromodulation, but the term is rarely used in this sense in the computational field of neuroevolution.}.
Although not explicitly stated, the dynamics of inhibition and excitation are also present in many machine learning methods such as Decision Trees (DT).
If we consider nodes in a DT to be like neurons (in neural trees the DT's nodes are exactly the same as neurons \cite{murthy1998automatic},\cite{golea1990growth}, \cite{sirat1990neural}), instead of processing information through some function and passing to the next connected neurons, they process information and chooses one of the connected neurons to activate.
Self-Organizing Map (SOM) is another example that uses excitation and inhibition.
SOM makes neurons compete for the input where only one will win and activate \cite{kohonen1990self}.
Both SOM and DT usually uses a crisp division, but a fuzzy division can easily be accomplished with fuzzy memberships in a net of excitation/inhibition neurons.

From an information processing perspective, the importance of using inhibition and excitation lies in the use of the divide-and-conquer approach to reduce complexity.
It is also possible to create network modules with inhibition and excitation as well as many other complex dynamics that are difficult to be created with other models.
Moreover, inhibition/excitation is a completely different model from the usual neuron model used by most neural networks. 
Therefore, it allows dynamics that are complex to build with the usual neuron model to be easily built. 

\subsection{Synaptic plasticity}

Synaptic plasticity has been motivated largely by its biological foundations, but few is known about its useful information processing properties.
There are homosynaptic and heterosynaptic mechanisms \cite{mouret2014artificial}:
\begin{itemize}
	\item Homosynaptic - weight update is only a function of the pre and post synaptic neurons (e.g., Hebbian learning);
	\item Heterosynaptic - the synaptic is neuromodulated, i.e., the same as homosynaptic with the addition of a modulatory neuron that influences strongly the Hebbian plasticity (e.g., modulated Hebbian learning).
\end{itemize}
In homosynaptic mechanisms, researchers reported that the use of synaptic plasticity aided their methods to adapt to environment changes \cite{urzelai2001evolution} but depending on the task it may not give any improvements \cite{stanley2003evolving}.
It seems that the change of weights based on the relationship between the connected neurons (e.g., Hebbian learning) creates certain complex dynamical systems as a result, but there is no reason why the same dynamical systems could not be created with a given structure and some neurons, i.e., without changing weights online.

\begin{figure}[ht]
 \centering
 \includegraphics[width=0.4\textwidth]{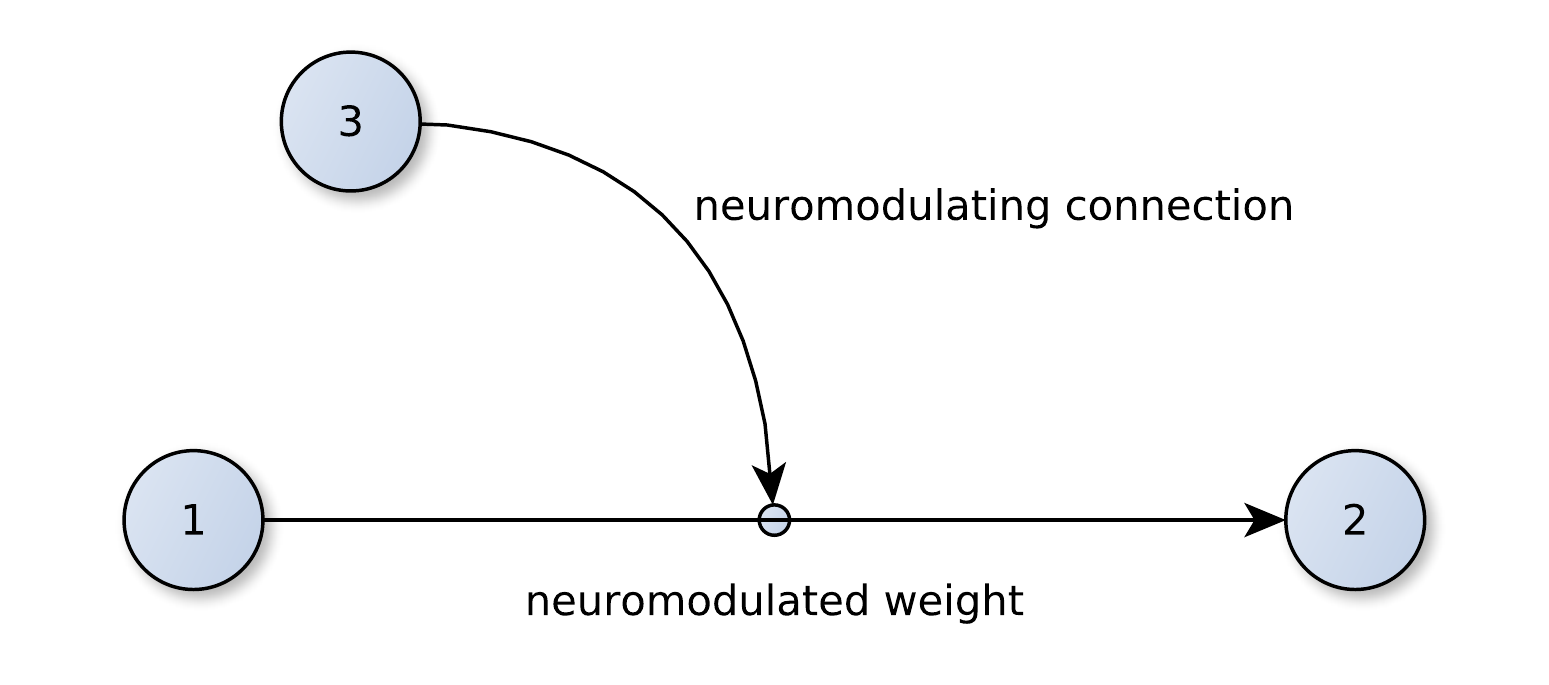}
 \caption{Given two neurons $1$ and $2$, the modification of their connection by another neuron is called neuromodulation of connections. To avoid confusions with other inhibition/activation types of neuromodulation, the diagram used here differs from the usual ones found in neuroevolution papers.}
 \label{nm2}
\end{figure}

Having said that, heterosynaptic mechanisms employ neuromodulation\footnote{Neuromodulation, in artificial neural network, is the modulation of connection weight(s) by another neuron.} of the connections and differ substantially from homosynaptic mechanisms.
The results obtained from using such mechanisms are much more encouraging \cite{soltoggio2007evolving}, \cite{soltoggio2008evolutionary}, \cite{risi2009novelty}, \cite{niv2002evolution}, \cite{kondo2007evolutionary}.
From an information processing point of view, the neuromodulation allows for another neuron to influence the relationship between two connected neurons.
Recall that the relationship between neurons coded by the connection is never dependent on another neuron.
Therefore, the heterosynaptic mechanism expands arguably the representation capabilities of the computation model.
This work will use only heterosynaptic mechanisms. 
Figure~\ref{nm2} shows a simple example.

\subsection{Feedback}

\begin{figure}[ht]
 \centering
 \includegraphics[width=0.13\textwidth]{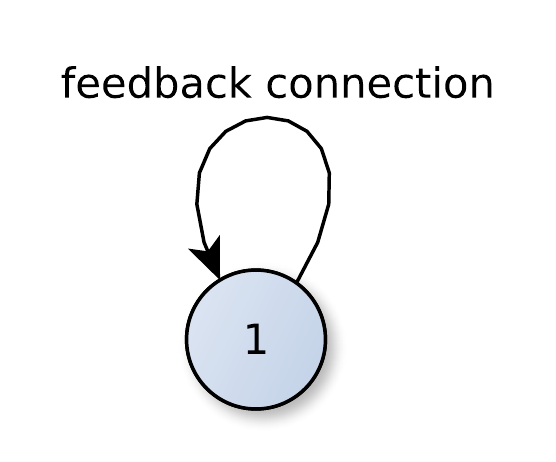}\\
 \includegraphics[width=0.4\textwidth]{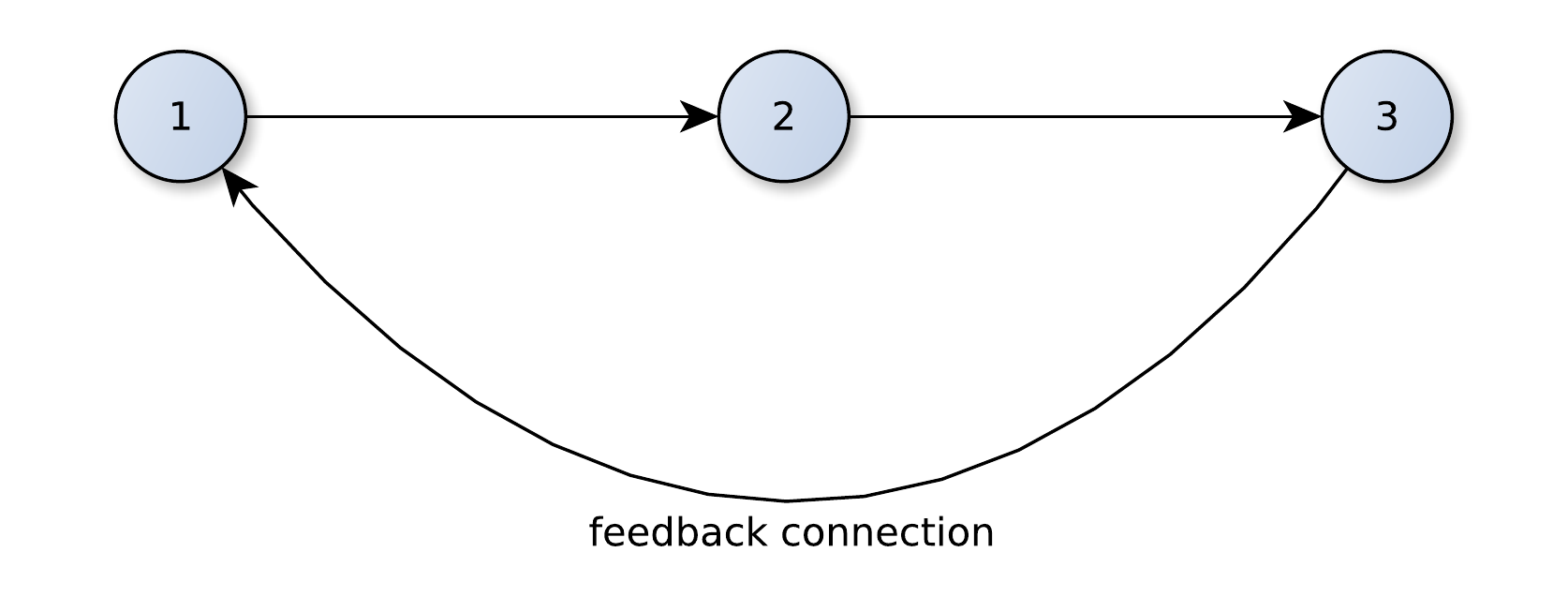}
 \caption{Two examples of neuron feedback.}
 \label{feedback1}
\end{figure}

Feedback is one of the ways that a network has to keep information from previous runs (e.g., one way of codifying memory).
This is the reason why neural networks with feedback can solve non-Markov problems.
Most importantly it is the only way that neurons in a processing sequence may have access to the resulting outputs from subsequent neurons (see Figure~\ref{feedback1} for examples of network structures with feedback).
This property is extensively used in control systems, electronic engineering, biology, social sciences and many other areas.

\section{Unified Neural Model}

In this section, the proposed unified neural model will be described.

\subsection{The Extended Neuron}

Neural models are built from the interconnections and features of simple units, the neurons.
The neuron used here is an extended version of the widely used neuron model employed by the perceptron, multi-layer perceptron, and most other neural network algorithms (will be referred to as "usual neuron model" in this paper). 
This extension is done in two ways:
\begin{itemize}
\item Adaptation Speed - Every neuron has its own reacting speed, that is, some neurons may take longer to change their outputs;
\item Type - Neurons have various activation functions. %They are described in detail in Table~\ref{table_activation_functions}.
\end{itemize}

%\begin{table*}[!ht]
%\centering
%\caption{Activation Functions}
%\begin{tabular}{ |l|l| }
	%\hline
	%\multicolumn{3}{ |c| }{Team sheet} \\
%	\hline
	% & GK & Paul Robinson \\ \hline
%	Name & Equation \\ \hline
%	Random & uniform random $\in (-1,1)$ \\
        %Threshold & $\left\{\begin{array}{cc} 1 & if $input >= 0;\\
 	%-1 & otherwise, \\
	%\end{array} \right$
%        Identity & $input$ \\
%	Sigmoid & hyperbolic tangent$(input)$ \\
%	\hline
%\end{tabular}
%\label{table_activation_functions}
%\end{table*}

\begin{figure*}[!ht]
\centering
 \includegraphics[width=0.8\textwidth]{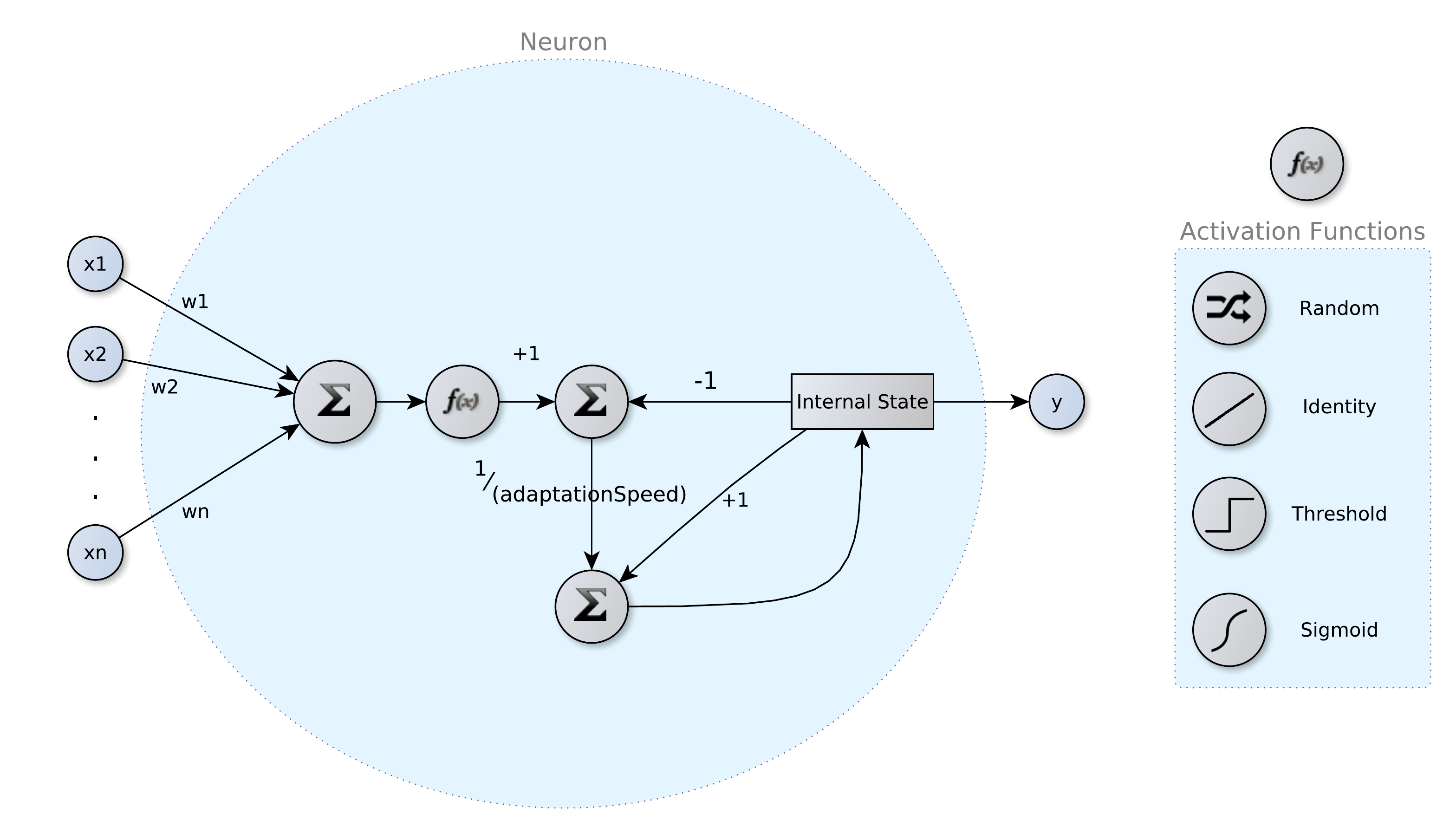}
 \caption{Neuron activation process and the possible activation functions.}
 \label{neuron_definition}
\end{figure*}

Figure~\ref{neuron_definition} shows a diagram of the behavior of a single neuron. 
Mathematically speaking, it can be determined by the following equations:
\begin{equation}
	a=f(\sum_{i=1}^{n}(w_i x_i))
\label{usual_perceptron_equation}
\end{equation}
\begin{equation}
\begin{split}
	&Ins_{t}= Ins_{t-1}+\frac{1}{adaptationSpeed}(a - Ins_{t-1})\\
	&y = Ins_{t}
\end{split}
\label{delta_rule}
\end{equation}
%\begin{equnarray}
%\end{equnarray}
%\end{equation}
Equation~\ref{usual_perceptron_equation} is basically the same as the usual neuron model, where every input $x_i$ is multiplied by its associated weight $w_i$ for $n$ inputs.
The result $a$ from applying the activation function $f()$ is, however, not used to update the output $y$ directly.
It is used instead in a delta rule like equation (Equation~\ref{delta_rule}) that takes the neuron adaptation speed $adaptationSpeed$ as parameter to update its previous internal state $Ins_{t-1}$.
The reasoning behind this equation is that if the neuron input changes, the neuron output would change completely after $adaptationSpeed$ iterations, i.e. creating a slower neuron.
Finally, the current internal state $Ins_{t}$ is used as output.

The reason behind these modifications go beyond bio-inspiration.
Different adaptation speeds result in a multi-scale network that is capable of taking long-term changes into account.
For example, finding an approximation to the average of a variable is as simple as connecting a slower neuron to it.
Moreover, the use of a single type of neuron (only one activation function) is an unnecessary bias.
The use of some types of neurons allows for the neural network to fit different problems without increasing much the searching space.

\subsection{Control Neuron and Control Signals}
\label{control_neuron_section}

One of the motivations behind TWEANNs is to evolve problem-related structures automatically.
In fact, sometimes a simple structure that switches between different behaviors (set of neurons) is enough to solve a task \cite{lessin2013open}.
However, there is no way to turn off a set of neurons in current TWEANNs so a question is raised.
How can structures such as a basic switch system come to life?

%To be able to divide the problem, structures that switch on and off other structures need to be present.
We propose to solve this problem by adding a different signal type called control signal.
Control signals will be solely responsible for activating and deactivating neurons, i.e. they will not interfere with their inputs.
%These systems can even evolve a self-organizing map.
Therefore, the following equation is added to decide whether the extended neuron activates or not:
\begin{equation}
\begin{split}
	&stimulation=\sum_{i=0}^{n}(w_i cs_i)\\
	&activation = \left\{\begin{array}{cc}
 true & \mbox{if $stimulation >= threshold$};\\
 false& \mbox{otherwise},
\end{array} \right. 
\label{control_signal}
\end{split}
\end{equation}
where $cs_i$ is the control signal and $w_i$ its relative weight for $n$ control signals.
Therefore, the neuron activates if the $stimulation$ is more or equal to the parameter $threshold$, because the $activation$ is set to $true$.
Otherwise, the neuron does not activate.
Here, the $threshold$ is fixed for all neurons.

To make all these possible, control signals need to be somehow present.
This makes necessary the introduction of an special neuron called control neuron.
The control neuron is exactly like all other neurons aside from the fact that it does not output normal signals but control signals.
All control neurons have threshold activation functions.

\subsection{The Genome}

\begin{figure*}[!ht]
\centering
 \includegraphics[width=1\textwidth]{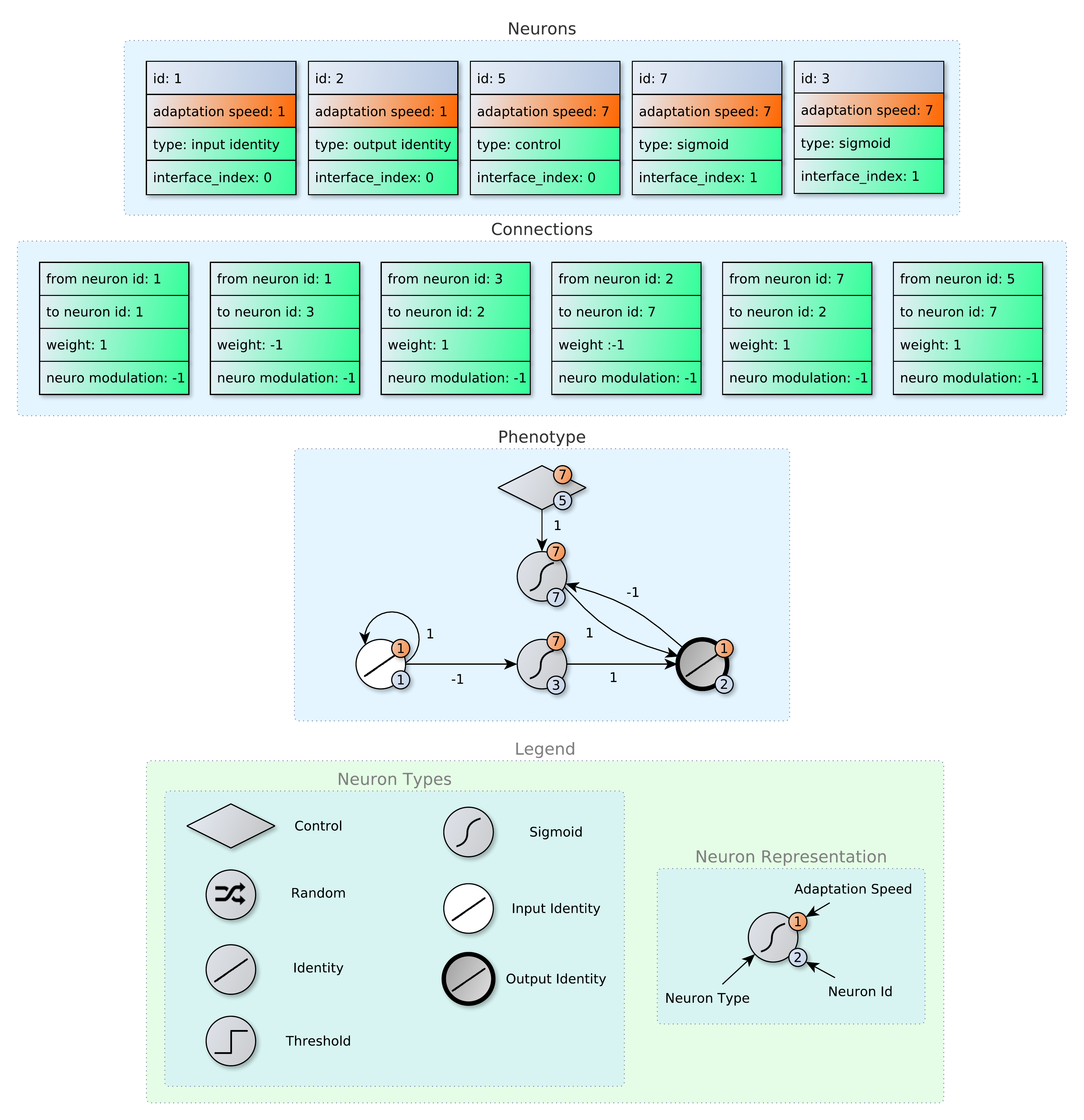}
 \caption{An example demonstrating how the genome maps its phenotype.}
 \label{genome}
\end{figure*}

To evolve neural networks it is necessary to define a genome.
The genome used is a direct encoding scheme made by a list of connections and a list of neurons (see Figure~\ref{genome}).

The evolvable parameters and their respective range values are shown in Table~\ref{para_range}, where $N$ is the set of neurons present in the respective chromosome.
Notice that the type parameter values {\em input identity} (input neuron with identity activation function) and {\em output identity} (output neuron with identity activation function) (see Figure~\ref{genome}) as well as the whole interface index parameters are not part of the evolvable parameters although they are present inside the genome.
This happens because they are only created by the initializing step of the evolution and kept fixed afterwards.

%\begin{equation}
%\begin{split}
%	Neuron\\
%	&type={}\\
%	&firingrate={}\\
%\label{control_signal}
%\end{split}
%\begin{aligned}
%	Neuron\\
%	Weight\\
%\end{aligned}
%\end{equation}

\begin{table*}[!ht]
\centering
\caption{Evolvable Parameter Space}
\begin{tabular}{ |c|l|l| }
	%\hline
	%\multicolumn{3}{ |c| }{Team sheet} \\
	\hline
	% & GK & Paul Robinson \\ \hline
	Related to & Parameter & Range \\ \hline
	\multirow{2}{*}{Neuron} & $adaptationSpeed$ & $\{1,7,49\}$ \\
	& Type  & $\{control,random,sigmoid,threshold,identity\}$ \\\hline
	\multirow{2}{*}{Weight} & From Neuron & $\{N\}$ \\
        & To Neuron & $\{N\}$ \\
        & Weight & $\{-1,1\}$ \\
        & Neuromodulation & $\{-1,N\}$ \\
	\hline
\end{tabular}
\label{para_range}
\end{table*}

%\begin{figure}[ht]
%\centering
% \includegraphics[width=0.5\textwidth]{diagrams/neuron_types.pdf}
% \caption{The various neuron types and the full neuron representation.}
% \label{nm1}
%\end{figure}

\subsection{Order of Neuron Execution}

To let the control neurons decide which neurons are going to execute, the control neurons need to be executed first.
Afterwards, control signals are established and the remaining neurons are executed.
In other words, the following order is respected:
\begin{itemize}
	\item Input neurons;
	\item Control neurons that do not receive connections from other control neurons;
	\item Remaining control neurons that are activated;
	\item Remaining neurons that are activated and have an absolute output higher than $0.001$;
\end{itemize}		

Input and output are always cleared after the network execution even when the respective neurons were not activated.
That is, if a given input or output neuron was not activated, the respective input or output is set to zero.

\section{Evolving the Unified Neural Model}

To evolve structures, special evolutionary methods are needed.
A new evolutionary method is proposed that is based on a new diversity paradigm.
Here, we show how diversity of structures can be achieved without creating a dependence on the problem.

\begin{figure*}[!ht]
\centering
 \includegraphics[width=0.8\textwidth]{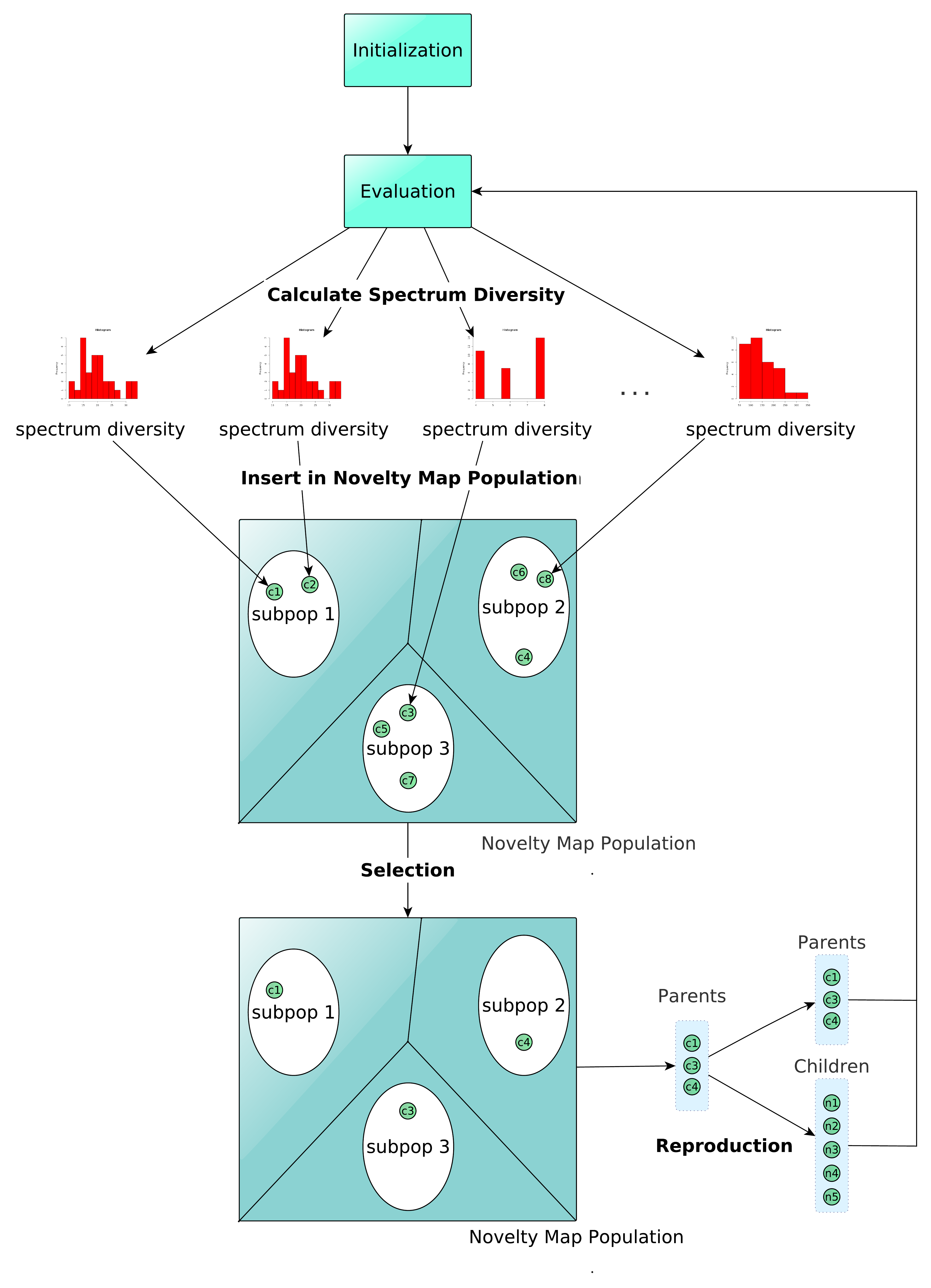}
 \caption{Evolution's overview.}
 \label{evolution}
\end{figure*}

An overview of the evolution process is shown in Figure~\ref{evolution}.
The steps will be described in detail in the following subsections.

\subsection{Initialization}

First, all chromosomes in the population are initialized to be a set of  input and output neurons, one for each input/output present in the problem.
For these neurons the interface index is set to be the index of the respective input/output.  
%Notice that they begin without any connections.
To create an initial diversity, some initial mutations are applied for all chromosomes.

\subsection{Evaluation}

In the evaluation stage every chromosome present in the population is activated for a whole trial.
At the end of each trial, the average reward received throughout is used as fitness.
This process is repeated until the entire population has been activated.
%A trial is defined as set of steps for the multi-step problem to terminate
%Afterwards they are inserted into the Novelty Population Map. 
Afterwards, the spectrum of every chromosome is calculated.

\subsection{Spectrum Diversity}

Diversity and protection of innovation is a challenging problem for TWEANNs, basically, because finding a metric to compare the structures of TWEANNs is difficult.
Speciation tries to approximate this comparison by using innovation numbers (a certain type of id) on the genes and therefore the alignment of such genes gives a fast approximation of the structure similarities \cite{stanley2002evolving}.
However, two structures originated from different evolutionary paths are always seen as totally different and unmatching structures because they differ in innovation numbers.
This happens even if they are the same or similar in structure.
And since genes are compared one per one it does not scale well with the size of the chromosome.

In Physics and Chemistry, the opposite problem takes place. 
Atoms combine with each other into complex structures, but it is discovering these structures which is a hurdle.
%When comparing objects such as plants or planets, not just the structures, but also the number of atoms make comparison difficult.
To solve these problems various types of spectra are used (e.g., nuclear magnetic resonance spectrum).

Similarly, here we propose to use spectra to identify and compare structures.
The spectrum is itself designed for the representation though, composed of a set of genotype's and phenotype's properties.
Computationally speaking, the spectrum is an array with each property being an element of this array.
Here the spectrum is composed of the following properties:
\begin{itemize}
	\item The number of identity neurons;
	\item The number of sigmoid neurons;
	\item The number of threshold neurons;
	\item The number of random neurons;
	\item The number of control neurons;
	\item The number of slower neurons (adaptation speed greater than one).
\end{itemize}
This combination of properties divides the set of solutions by how they approach to solve the problem.

Once the spectrum of all chromosomes are calculated, their spectrum are fed into the Novelty Map \cite{vargas2015novelty} (Section~\ref{novelty_map_section}) where only the spectra that are mapped into the same cell creates a group (species) and compete.
The entire process of calculating the spectra and insertion in the Novelty Map is called Spectrum Diversity, because the spectrum of the chromosomes is used to separate the population into groups(species), i.e., diversity is kept by creating species based on their spectrum similarity.

Spectrum diversity uses a measure of diversity based on the distance between the spectra of two structures.
The advantages of using this type of diversity are:
\begin{itemize}
	\item Problem Independence - The metric is not defined based on the problem characteristics, therefore it works independent of problem type and size;
	\item Scalability - Since the spectrum should describe properties of the structure as a whole, the size of the chromosome does not interfere with the metric.
\end{itemize}		

Notice that measures from complex networks and graph theory can be also be used, joining the discipline of complex networks with TWEANNs.
Actually, by using metrics from the complex networks field it is possible to achieve many types of structure diversity, for example, degree distribution diversity, assortativity diversity, etc. 

\subsection{Novelty Map}
\label{novelty_map_section}

Novelty Map is a table with the most novel inputs according to a given novelty metric (see Figure~\ref{nmap_alg}).
When an input is presented to the map, competition takes place where the cell with the closest weight array wins.
This winner cell is activated and can be used in many ways, here the winner cell is used to identify to which species (group) the input pertains.
Afterwards, the table is updated by substituting the weight array of the least novel cell (according to the novelty measure) with the input array if and only if the input array has higher novelty.
%This way the table is always kept up to date.

\begin{figure}[!ht]
\centering
\begin{algorithmic}[1]
%Parameters:
%\begin{enumerate}
%\item $Max_n$: maximum size of the map;
%\item Novelty metric;
%\end{enumerate}
	\REQUIRE $|P| > Max_n > 0$ and $Nmetric() \neq \emptyset$
%	\STATE $|NM|=0$ \COMMENT{Set the size of the map to zero}
	\STATE $NM \leftarrow \emptyset$ \COMMENT{All cells are empty}
	\LOOP
	\IF {An input $I$ is presented to the Novelty Map}
	%\IF [if there is an empty cell]	{$|NM| < Max_n$}
	\IF {$|NM| < Max_n$}
	\STATE $NM \leftarrow NM\cup\{I\} $ %Insert the input in an empty cell of the map
%	\STATE $n=n+1$ (Increment the size of the novelty map)
	\ELSE 
%	\STATE $Nmetric(I)$ Evaluate the input's novelty with the novelty metric
	\STATE $\{A \in NM~|~\forall K \in NM \land Nmetric(A)<Nmetric(K)$\}
	\IF {$Nmetric(I) > Nmetric(A)$}
	\STATE $NM \leftarrow NM\cup\{I\} $ %Insert the input
	\STATE $NM \leftarrow NM\setminus\{A\}$ %Remove the sample with the lowest novelty from the map
	\ENDIF
	\ENDIF
	\RETURN $H:\{H \in NM~|~\forall K \in NM \land dist(I,H) < dist(I,K)\}$ \COMMENT{Return cell which is closest to input $I$}
	\ENDIF
	\ENDLOOP
 \label{nmap_alg}
\end{algorithmic}
\caption{Novelty Map algorithm. $P$ is the population size, $Nmetric()$ is a given novelty metric, $NM$ is the set of cells in the Novelty Map, $Max_n$ is the maximum size pf $NM$ and $dist()$ is the Euclidean distance.}
\end{figure}

%\begin{table}[!h]
% \centering
%\caption{Novelty Map Algorithm} 
%\begin{tabular}{p{8cm}}
%\hline
%Parameters:
%\begin{enumerate}
%	\item $Max_n$: maximum size of the map;
%	\item Novelty metric;
%\end{enumerate}
%Set the size of the map $n$ to zero ($n=0$)\\
%All cells are empty\\
%Infinite Loop:
%\begin{enumerate}
%\item When an input is presented to the novelty map do:
%\item If $n < Max_n$ (i.e., if there is an empty cell)
%\begin{enumerate}
%\item Insert the input in an empty cell of the map
%\item Increment the size of the novelty map ($n=n+1$)
%\end{enumerate}
%else
%\begin{enumerate}
%\item Evaluate the input's novelty with the novelty metric
%\item If the input's novelty is higher than the lowest novelty from the samples inside the map:
%\begin{enumerate}
%	\item Insert the input and remove the sample with the lowest novelty from the map
%\end{enumerate}
%\end{enumerate}
%\item Return the weight array of the cell which is closest to the input
%\end{enumerate} \\
%\hline
%\end{tabular}
%\label{nmap_alg}
%\end{table}

This table shares some similarities with the self-organizing map \cite{kohonen2001self} and the neural gas \cite{fritzke1995growing}, but its behavior is independent of input frequency and it uses fewer cells to map the same input space (cell's efficiency).

The novelty metric used in this article is the uniqueness.
Let $S$ be a set of arrays. 
Uniqueness is defined for an array $a_i$ in relation to the other arrays in $S$ by:
\begin{eqnarray}
	&U = S \setminus \{a_i\} \\
	&uniqueness = min_{a_k \in U}(dist(a_i,a_k)).
\end{eqnarray}
%In other words, uniqueness of an array is the smallest distance  to the respective array for any array present in the set, excluding the array itself.
%This novelty metric was chosen because of its simplicity and quality though any other novelty measure could have been used instead.

\subsubsection{Novelty Map Population}

Novelty Map Population is a Novelty Map where cells are subpopulations and inputs are some value related to the individual.
Here, the spectrum of chromosomes are used as input.

\subsection{Selection}

In the selection stage, only one individual survives inside every cell (species) of the Novelty Map.
To decide which individual is going to survive, the individuals are compared based on their fitness.
And in the case of draw, the individual that uses less neurons wins (i.e. the least complex wins).

In this sense, subpopulations are the same as niches, because chromosomes that fall inside the same subpopulation have a similar histogram and therefore are similar themselves.

After selection, the individuals that survived become parents and the algorithm enters in reproduction stage.
As a matter of fact, to keep the number of parents constant, if a subpopulation is empty, a random valid parent from a different subpopulation is chosen and copied instead.

\subsection{Reproduction}

With the parents defined, the remaining vacancies in the population are filled by the following sequence of procedures:
\begin{enumerate}
	\item A parent is chosen at random; 
	\item Some step mutations are applied to the parent to create a child;
	\item All connections have a $50\%$ chance of being perturbed with a random value in the range $[-w,w]$, where $w$ is the current weight of the respective connection.
\end{enumerate}		

A mutation can be any of the following procedures:
\begin{itemize}
	\item Add a neuron. Everything is randomly chosen, but first the neuron has an additional probability of being a control neuron. Additionally, two new connections are created connecting this neuron to the network (one random connection from this neuron and another random one to this neuron);
	\item Delete a neuron (all connection to the respective neuron are also deleted). Input/output neurons cannot be deleted;
	\item Add a connection. There is a probability of the connection be neuromodulated (neuromodulation probability). Everything else is randomly decided;
	\item Delete a connection.
\end{itemize}		
Any of the above mutation procedures have a chance of occurring defined by the mutation probability array.
The mutation probability array is composed of four real numbers relative to the probability of respectively adding a neuron, delete a neuron, add a connection and delete a connection.
So for example $\{0.1,0.1,0.4,0.4\}$ would mean $10\%$ chance of adding a neuron, $10\%$ chance of deleting a neuron, $40\%$ chance of adding a connection and $40\%$ chance of deleting a connection.

%Mutation probability array is composed of four real numbers, relative to the probability of 

\section{Experiments}

In this section both SUNA and NEAT will be tested on five different classes of problems, each one requiring different learning features to solve.
The set of parameters are kept the same for all the problems for both algorithms (see Section~\ref{exp_settings}).

\begin{table}[!ht]
\centering
\caption{Parameters for SUNA}
\begin{tabular}{ |l|l| }
	%\hline
	%\multicolumn{3}{ |c| }{Team sheet} \\
	\hline
	% & GK & Paul Robinson \\ \hline
	Parameter & Value \\ \hline
	  Number of initial mutations & $200$ \\
	  Number of step mutations & $5$ \\
	  Population size & $100$ \\
	  Maximum Novelty Map population size ($Max_n$) & $20$ \\
	  Mutation probability array & $\{0.01,0.01,0.49,0.49\}$ \\
	  Neuromodulation probability & $0.1$ \\
	  Control neuron probability & $0.2$ \\
	  Excitation threshold & $0.0$ \\
%	  Mutation probability & $0.0$ \\
	\hline
\end{tabular}
\label{para_notc}
\end{table}

\begin{table*}
\centering
\caption{Parameters for NEAT}
\begin{tabular}{ |c|l|c|l|c|l| }
	%\hline
	%\multicolumn{3}{ |c| }{Team sheet} \\
	\hline
	Parameter & Value & Parameter & Value \\
	\hline
	\verb|trait_param_mut_prob| & $0.5$ & \verb|trait_mutation_power| & $1.0$ \\
	\verb|linktrait_mut_sig| & $1.0$ & \verb|nodetrait_mut_sig| & $0.5$ \\
	\verb|weigh_mut_power| & $2.5$ & \verb|recur_prob| & $0.00$ \\
	\verb|disjoint_coeff| & $1.0$ & \verb|excess_coeff| & $1.0$ \\
	\verb|mutdiff_coeff| & $0.4$ & \verb|compat_thresh| & $3.0$ \\
	\verb|age_significance| & $1.0$ & \verb|survival_thresh| & $0.20$ \\
	\verb|mutate_only_prob| & $0.25$ & \verb|mutate_random_trait_prob| & $0.1$ \\
	\verb|mutate_link_trait_prob| & $0.1$ &	\verb|mutate_node_trait_prob| & $0.1$ \\
	\verb|mutate_link_weights_prob| & $0.9$ & \verb|mutate_toggle_enable_prob| & $0.00$ \\
	\verb|mutate_gene_reenable_prob| & $0.000$ & \verb|mutate_add_node_prob| & $0.03$ \\ 
	\verb|mutate_add_link_prob| & $0.05$ & \verb|interspecies_mate_rate| & $0.001$ \\
	\verb|mate_multipoint_prob| & $0.6$ & \verb|mate_multipoint_avg_prob| & $0.4$ \\
	\verb|mate_singlepoint_prob| & $0.0$ & \verb|mate_only_prob| & $0.2$ \\
	\verb|recur_only_prob| & $0.0$ & \verb|pop_size| & $100$ \\
	\verb|dropoff_age| & $15$ & \verb|newlink_tries| & $20$ \\
	\verb|print_every| & $5$ & \verb|babies_stolen| & $0$ \\
	\verb|num_runs| & $1$ & &\\

%	\verb|mutate_only_prob| & $0.25$ & \verb|mutate_random_trait_prob| & $0.1$ & \verb|mutate_link_trait_prob| & $0.1$\\
%	\verb|mutate_node_trait_prob| & $0.1$ & \verb|mutate_link_weights_prob| & $0.9$ \verb|mutate_toggle_enable_prob| & $0.00$ \\
%	\verb|mutate_gene_reenable_prob| & $0.000$ & \verb|mutate_add_node_prob| & $0.03$ & \verb|mutate_add_link_prob| & $0.05$ \\
%	\verb|interspecies_mate_rate| & $0.001$ & \verb|mate_multipoint_prob| & $0.6$ & \verb|mate_multipoint_avg_prob| & $0.4$ \\
%	\verb|mate_singlepoint_prob| & $0.0$ & \verb|mate_only_prob| & $0.2$ & \verb|recur_only_prob| & $0.0$ \\
%	\verb|pop_size| & $100$ & \verb|dropoff_age| & $15$ & \verb|newlink_tries| & $20$ \\
%	\verb|print_every| & $5$ \verb|babies_stolen| & $0$ & \verb|num_runs| & $1$ \\

	% & GK & Paul Robinson \\ \hline
	\hline
\end{tabular}
\label{para_neat}
\end{table*}

\subsection{Experiments' Settings}
\label{exp_settings}

The NEAT code used is the 1.2.1 version of the NEAT C++ software package \cite{neatcode}.
%Notice that although with different problem's settings (the initial position was randomized),  NEAT was previously applied to a discrete action Mountain Car \cite{whiteson2006evolutionary}.
%Therefore, both the settings used in that paper and the settings present in the original package were evaluated.
%In the end, the original package settings had better results, therefore the settings used for NEAT is the one provided with the software package (i.e. the same settings that was previously used in a double pole balancing task with success).
Table~\ref{para_neat} shows the NEAT parameters.
Actually, the used parameters are the same as the one provided with the package for solving double pole balancing task.
A couple of variations of it were tested, but the original one performed better.
Moreover, some modifications were necessary to make NEAT work on problems with negative fitness, since it does not work out of the box. 
Therefore, a value big enough ($2000$) was always added to the final accumulated fitness, transforming the negative fitness into a positive one.
Additionally, to satisfy NEAT's requirements, for all problems the input range was converted to $(-1,1)$ while the output range was transformed into $(0,1)$.
This conversion naturally informs the algorithm about the maximum/minimum of both input and output, simplifying the problems at hand.
Such changes are not necessary for the proposed method. 

Ablation tests are also conducted to evaluate the importance of most of its features as well as understand when they are important.
These tests are conducted over problems without any transformation of input/output.
In other words, the algorithm does not know the range of input/output, verifying further its learning capabilities.

All results are averaged over $30$ runs\footnote{Run is defined as the sequence of trials until the maximum number of trials is reached.} and only the best result among $100$ trials\footnote{Trial is a set of iterations between agent and environment until a problem defined stopping criteria is met.} is plotted.
Every run, when not specified otherwise, has a maximum of $2\times10^5$ trials.

\subsection{Mountain Car}

\begin{figure}[!ht]
\centering
\includegraphics[scale=0.3]{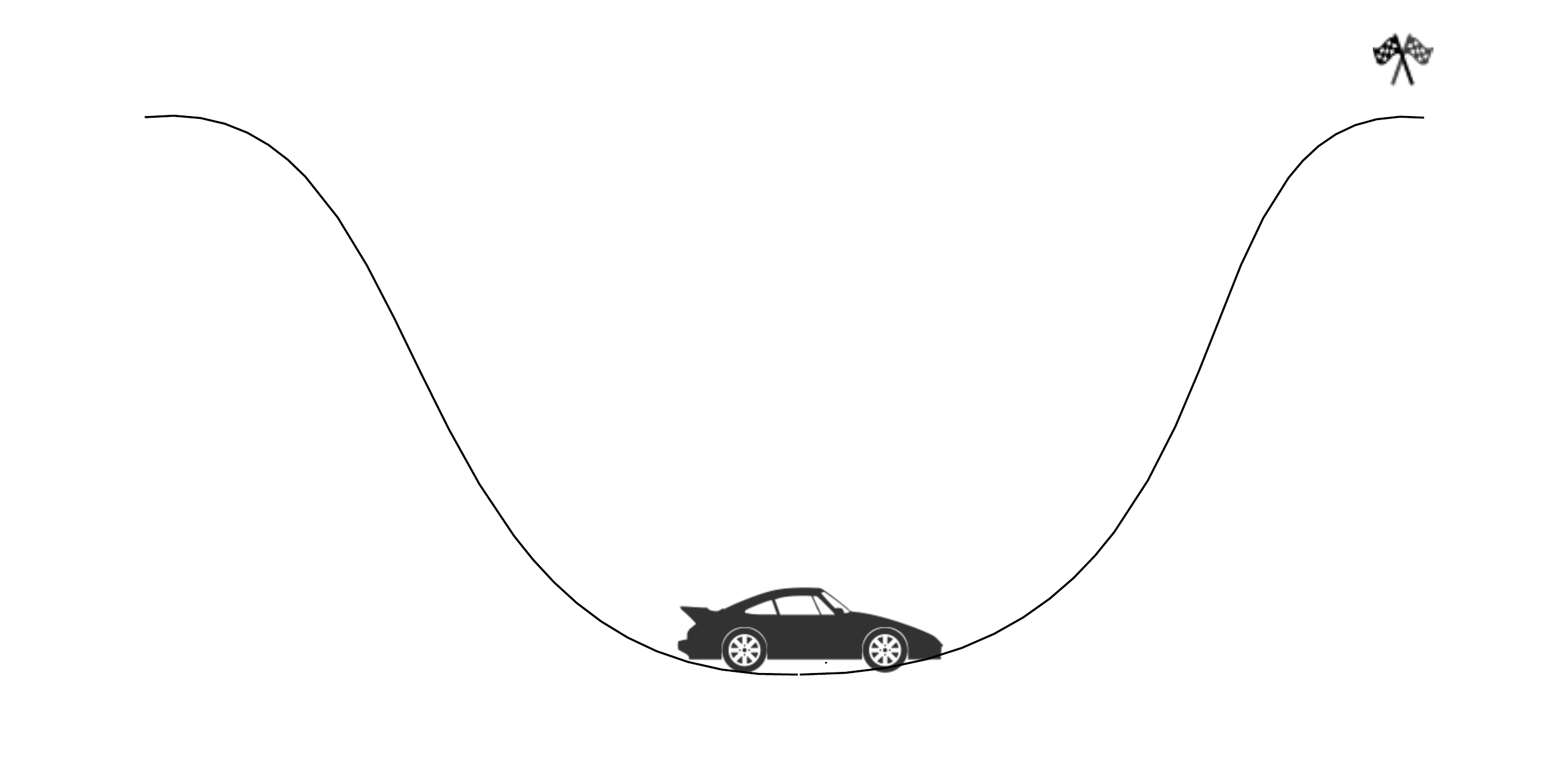}
\caption{Mountain car problem. The car's objective is to reach the flags uphill, although its acceleration is not enough to climb the mountain.}
\label{mountain_car}
\end{figure}

The mountain car problem \cite{sutton1996generalization} is defined by the following equations (Figure~\ref{mountain_car} shows an illustration of the problem):
\begin{equation}
\begin{aligned}
	&pos \in (-1.2,0.6) \\
	&v \in (-0.07,0.07) \\
	&a \in (-1,1) \\
	&v_{t+1} = v_{t} + (a_{t}) *0.001+\cos(3*pos_t)*(-0.0025) \\
	&pos_{t+1} = pos_t + v_{t+1},
\end{aligned}
\end{equation}
where $pos$ is the position of the car, $a$ is the car's action and $v$ is the velocity of the car.
The starting velocity and position are respectively $0.0$ and $-0.5$.
If $v < 0$ and $pos \le -1.2$, the velocity is set to zero.
When the car reaches $pos \ge 0.6$ the trial is terminated and the algorithm receives $0$ as reward.
In all other positions the algorithm receives $-1$ as a reward.
Moreover, if the algorithm's steps exceed $10^3$ the trial is terminated and the common reward of $-1$ is returned to the algorithm.

NEAT can only deal with certain ranges of input/output, therefore the input was converted to the $(-1,1)$ range while the output was converted to the $(0,1)$ range.
Naturally, this conversion informs the algorithm about the maximum/minimum of both input and output, an information that was not readily available in the original problem, therefore the problem is to some extent simplified.

\begin{figure}[!ht]
\centering
 \includegraphics[width=0.5\textwidth]{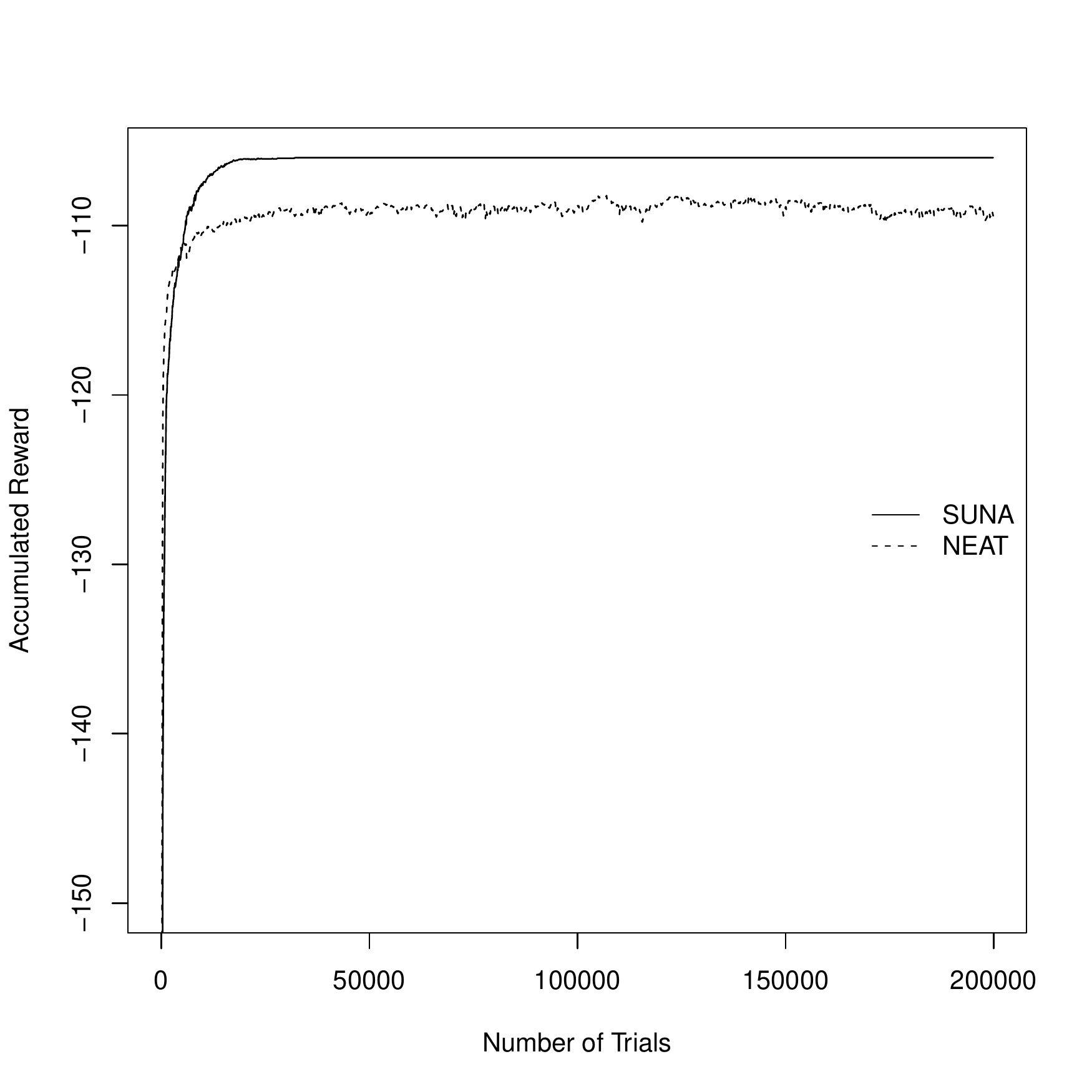}
 \caption{Comparison of SUNA and NEAT in the mountain car problem.}
 \label{mc_comparison}
\end{figure}

Figure~\ref{mc_comparison} shows a comparison between SUNA and NEAT in the mountain car problem.
SUNA performs better than NEAT, converging as fast as NEAT even though SUNA sees a bigger search space.
The reason behind the worse performance in NEAT seems to be related with unstable individuals taking over the population from time to time, making the results vary around the best solution but rarely staying in that position.
Moreover, procedures that check for stagnation make things worse when NEAT has already reached a good result.

\begin{figure}[!ht]
\centering
 \includegraphics[width=0.5\textwidth]{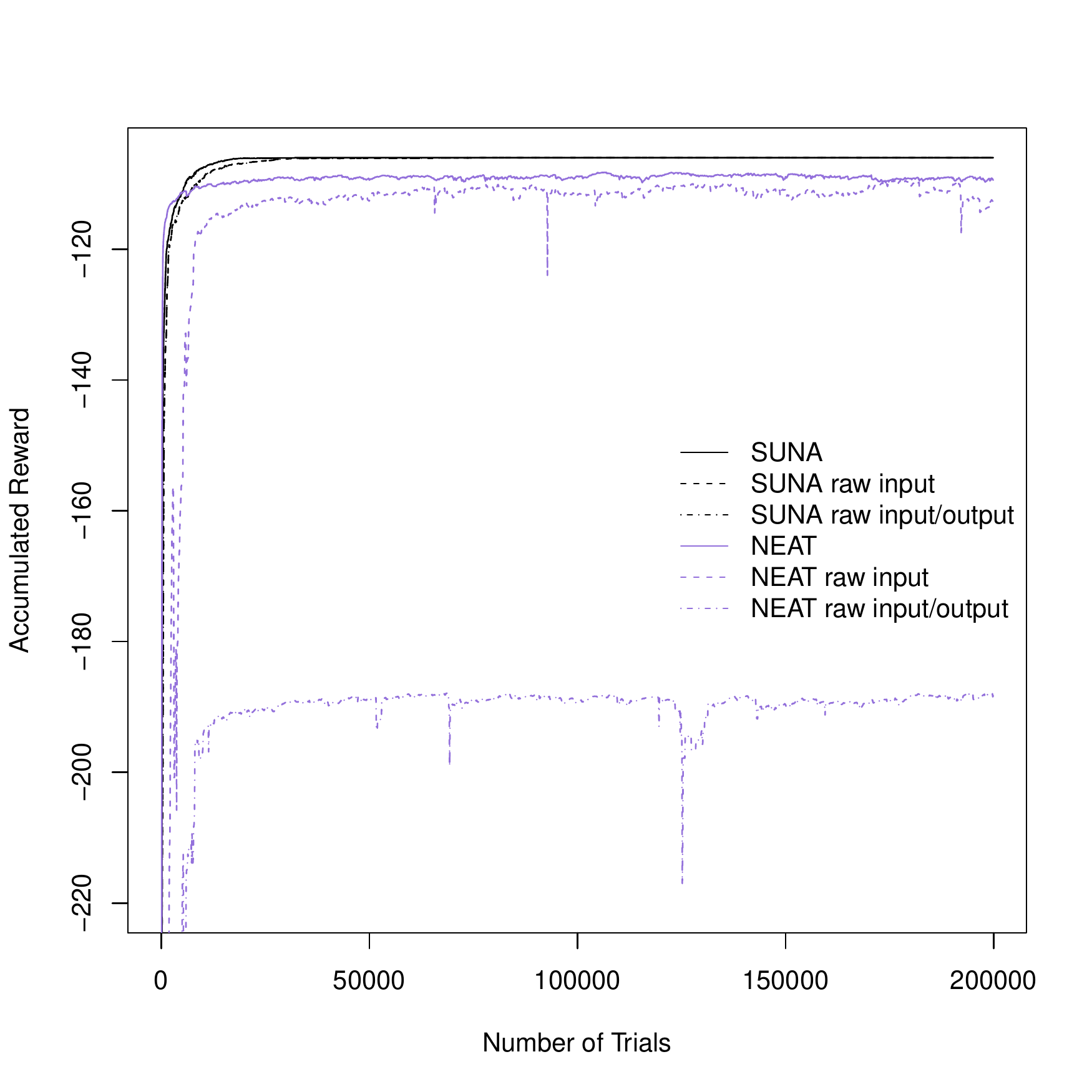}
 \caption{In real world problems, the maximum/minimum of either input or output is often not available, therefore a normalization is not readily possible. 
 Here tests show the performance of both SUNA and NEAT when both input and output are normalized (SUNA,NEAT), when the input is raw (SUNA raw input, NEAT raw input) or when both input and output are raw (SUNA raw input/output, NEAT raw input/output). 
 The tests were made in the mountain car problem.}
 \label{mc_raw}
\end{figure}

The non-simplified version of the problem, i.e., with raw input and output is shown in Figure~\ref{mc_raw}.
SUNA performs equally well on all problems, showing that the presence or absence of normalization in the input/output is not an issue.
NEAT, on the contrary, struggles with raw input and when the raw output is present it performs very poorly.
It is understandable, NEAT has only sigmoid input nodes that squeeze values into the $(-1,1)$ range and outputs nodes in the $(0,1)$ range and therefore can not cope with this type of problem.
Having said that, it is still a limitation, compromising the automatization and limiting the application to some problems.

\subsection{Double Pole Balancing}

\begin{figure}[!ht]
\centering
 \includegraphics[width=0.45\textwidth]{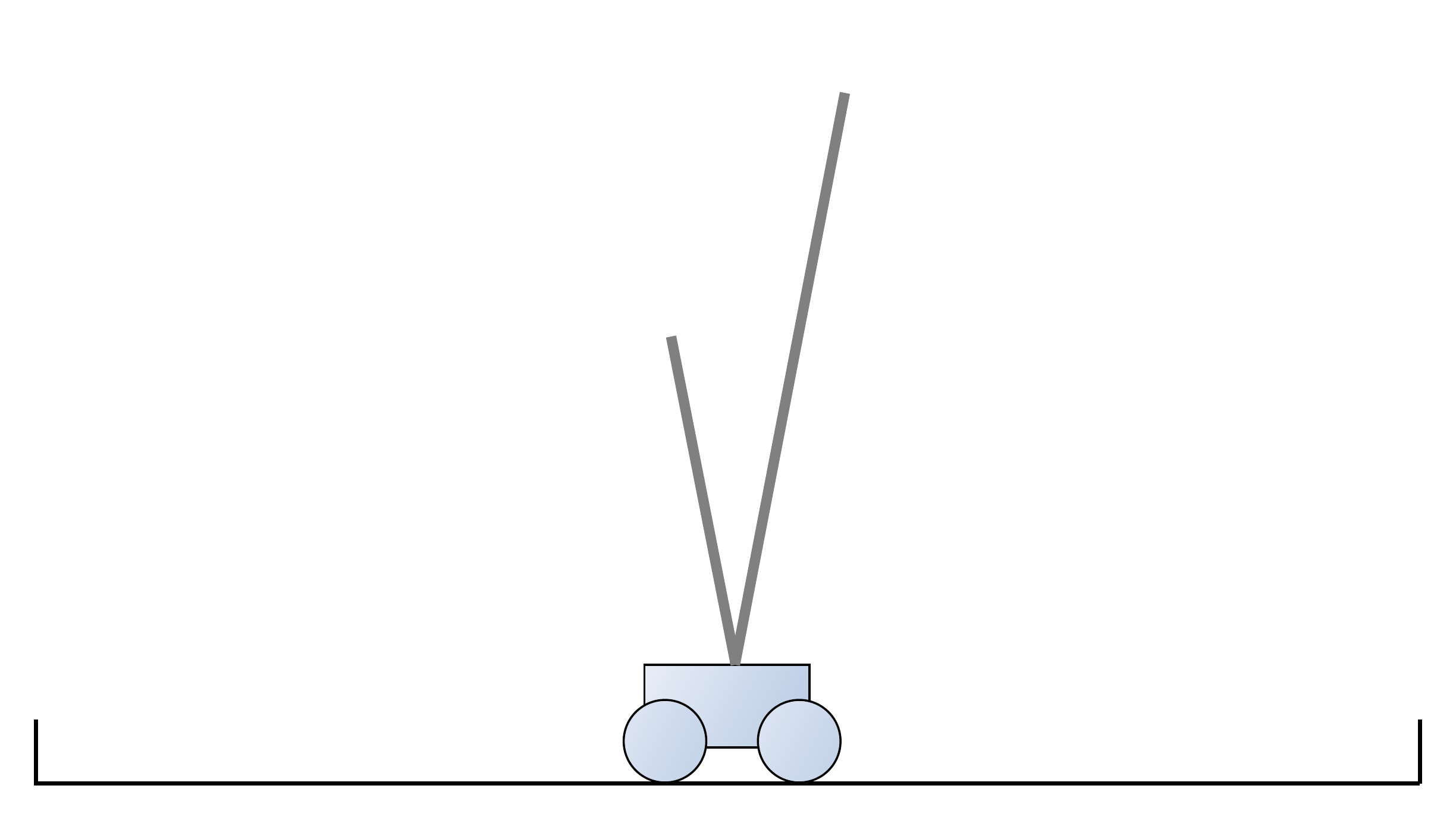}
 \caption{An illustration of the double pole balancing problem.}
 \label{doublepole}
\end{figure}

Double pole balancing is the famous problem of balancing a pole above a cart by only moving the cart forward or backward (see Figure~\ref{doublepole}).
This problem has six observing variables.
The dynamics used here are the same mentioned by the following papers \cite{wieland1991evolving},\cite{gomez1999solving},\cite{stanley2002evolving}. 
In fact, the code was adapted from \cite{neatcode} which is the same code used in other papers \cite{gomez1999solving},\cite{stanley2002evolving}.
All the original parameters of the problem were also kept the same.
%The objective function was changed to a simpler version.

\begin{figure}[!ht]
\centering
 \includegraphics[width=0.5\textwidth]{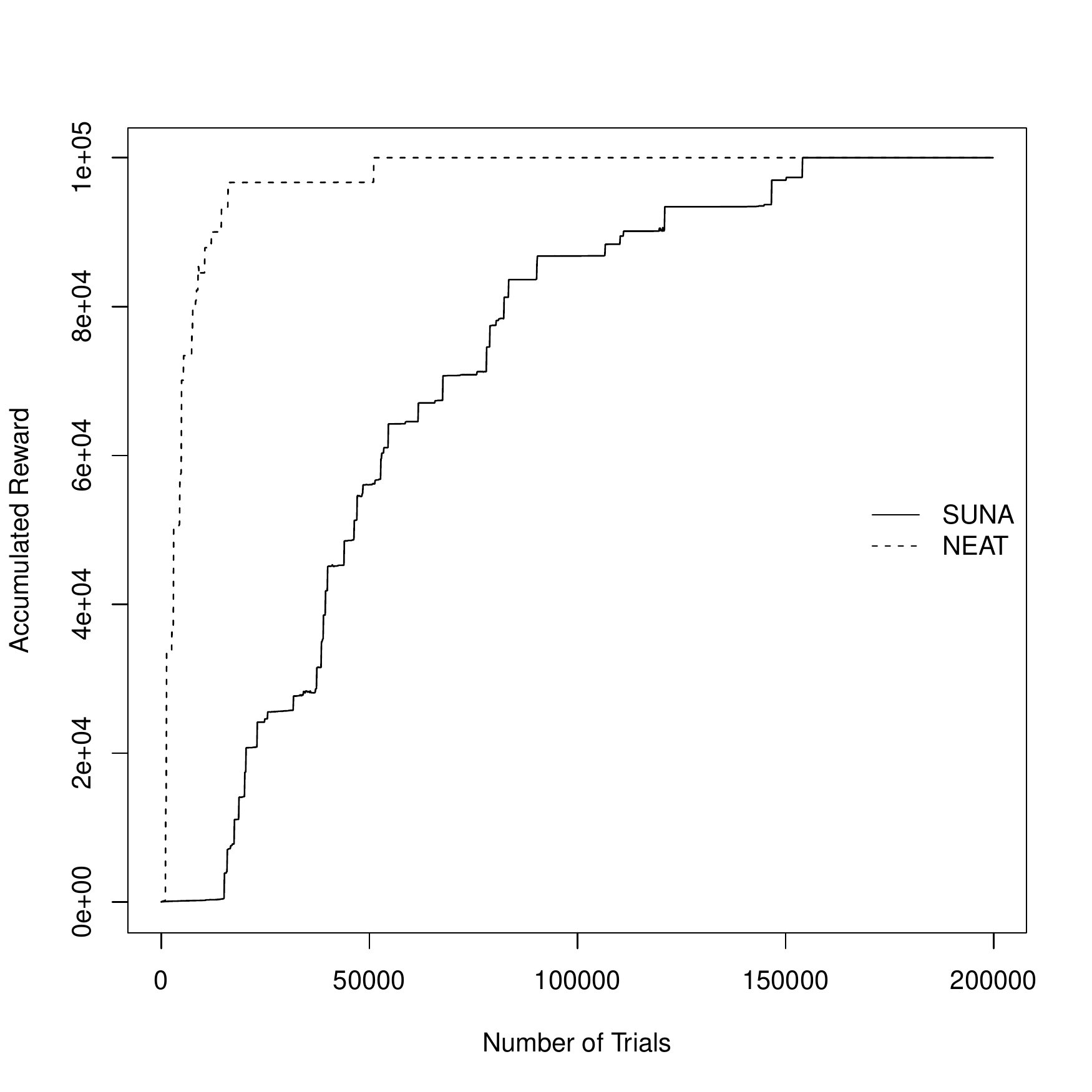}
 \caption{Double Pole Balancing.}
 \label{doublepole_results}
\end{figure}

Figure~\ref{doublepole_results} shows that SUNA and NEAT converges to the same final performance.
In other words, both algorithms learned to balance the pole for $10^5$ steps (the maximum allowed) in all the tests.
SUNA converges slower than NEAT.
This is expected due to the greater complexity of the SUNA's model.
%However, this is not a rule as we saw
%However, even with a bigger search space SUNA find an equally good solution.
%In this problem

\subsection{Non-Markov Double Pole Balancing}

Non-Markov double pole balancing has the same dynamics as the double pole balancing.
However, the agent only observes three variables (position of the cart and angles of both poles) instead of six.
To keep the poles balanced it is necessary to estimate the velocity of the cart and pole angles.
Therefore, this problem presents an additional difficulty.
The previous input needs to be somehow kept and used to compose an estimate of velocity.

Notice that in this article the double pole balancing fitness function is used, i.e., a modified fitness function that penalizes oscillations is not used \cite{stanley2002evolving}.
Thus, this formulation of non-Markov double pole balancing problem is relatively more difficult.

\begin{figure}[!ht]
\centering
 \includegraphics[width=0.5\textwidth]{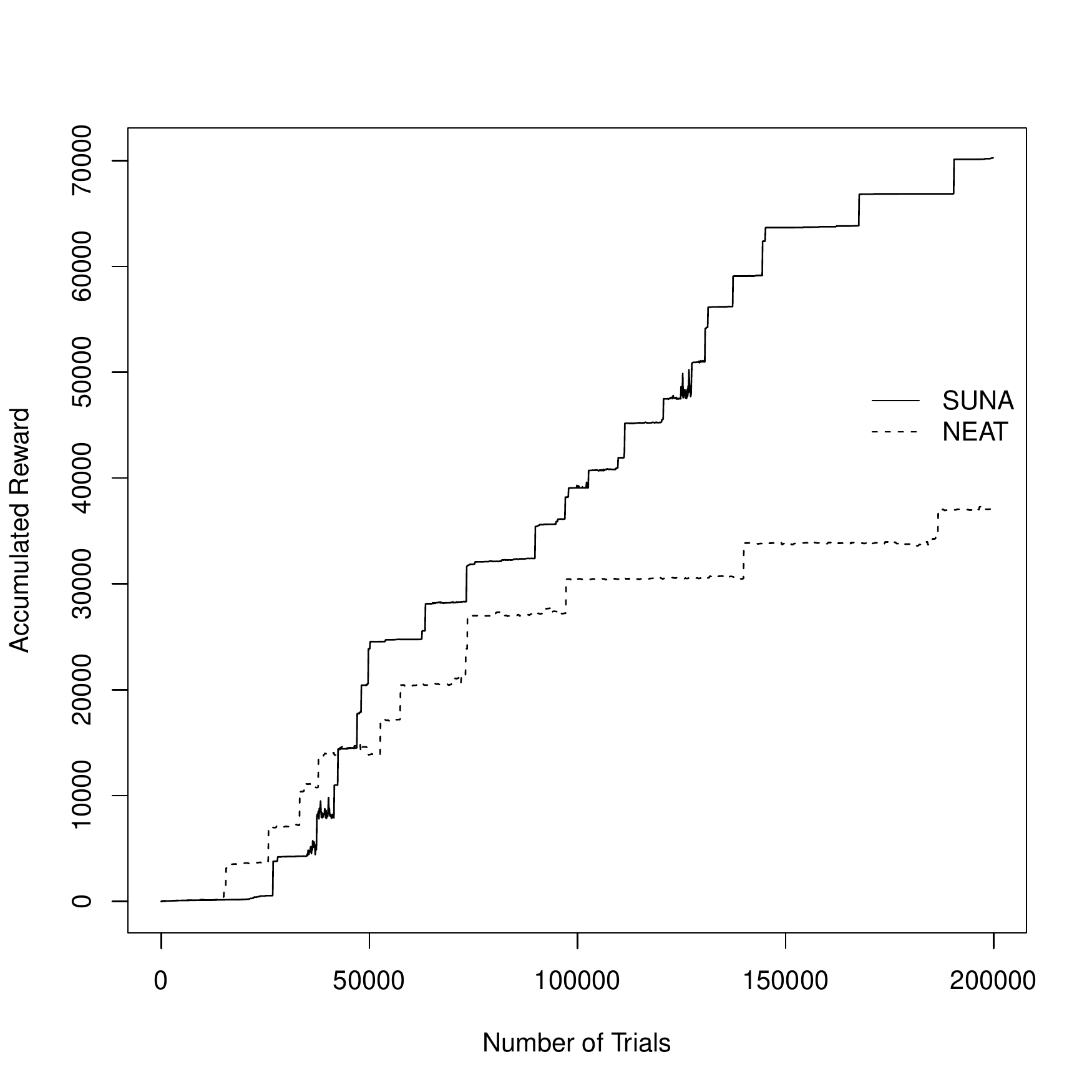}
 \caption{Performance of both SUNA and NEAT in the non-Markov double pole balancing problem.}
 \label{nmdp_res}
\end{figure}

Comparison results are shown in Figure~\ref{nmdp_res}
SUNA performs much better than NEAT in this problem.
In fact, SUNA achieves $70\%$ of the maximum possible average accumulated reward while NEAT reaches only $37\%$.
But what features present in the proposed algorithm enable it to outperform NEAT?
This question is answered in Section~\ref{ablation_tests} with ablation tests.
However, there are basically three important features: slow neurons (responsible for computing approximation of derivatives, averages and so on), real weights (accurate computation of estimates) and random neurons (exploration).

\subsection{Multiplexer}

Multiplexer is a binary problem with a single bit output and with the input composed of data and address variables.
To compute the correct output, $abits$ address variables are used to select one of the other $2^{abits}$ data variables.
This selected data variable is the correct output.
%In this manner, a given learning algorithms needs to decide based on the address and data inputs which binary value to output.
In this manner, the algorithm must first separate the address input from the data input and then use the address information to select the data part of the input.
This problem is coded as a reinforcement learning problem where each correct answer is rewarded with $0$ and each incorrect answer is rewarded with $-1$.
Every possible set of inputs are presented during a trial in random order, not allowing learning algorithms to memorize the sequence of outputs instead of the multiplexer function.

\begin{figure}[!ht]
\centering
 \includegraphics[width=0.5\textwidth]{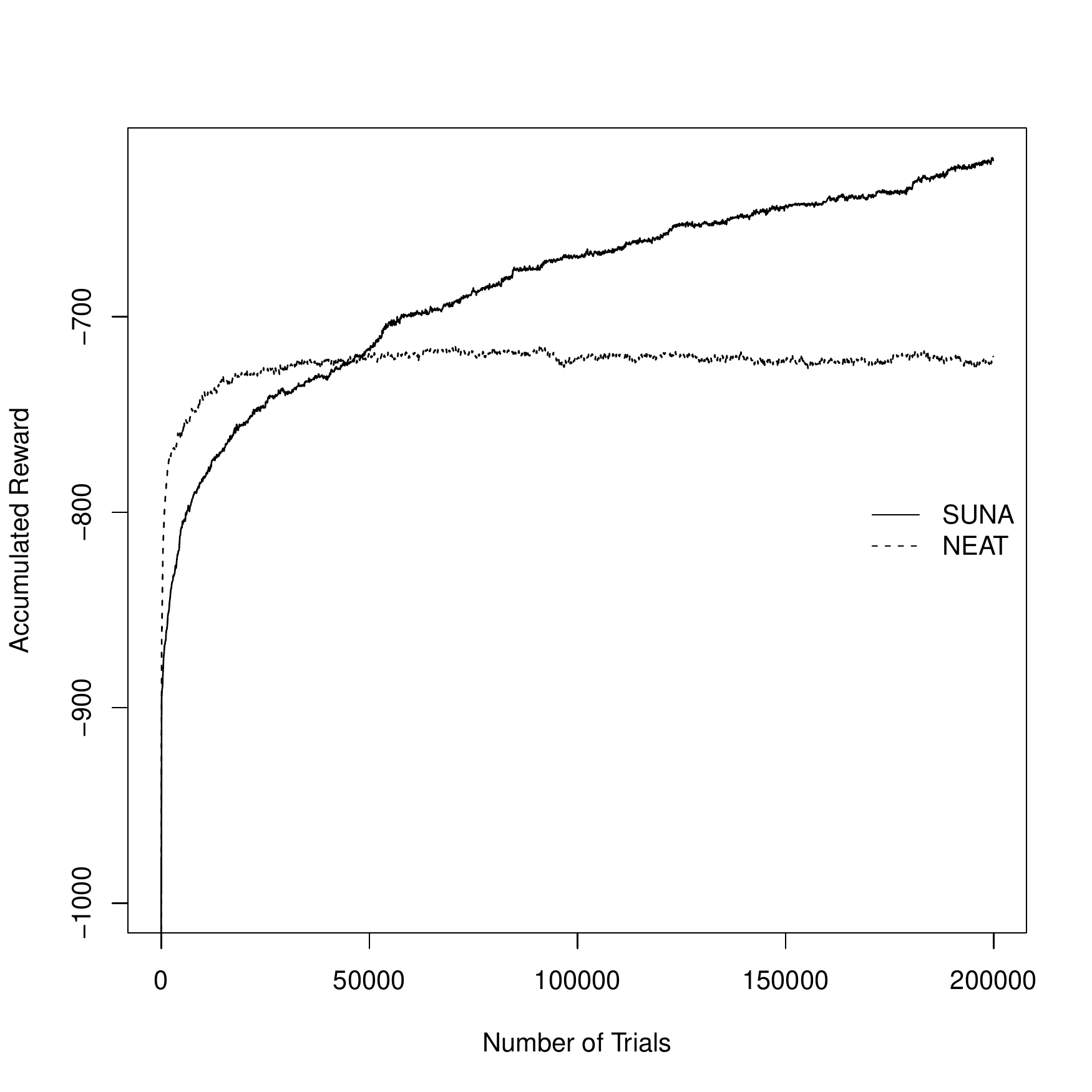}
 \caption{SUNA outperforms NEAT in the multiplexer problem.}
 \label{multiplexer_results}
\end{figure}

Figure~\ref{multiplexer_results} shows that SUNA outperforms NEAT in the multiplexer problem with $abits=3$ (i.e., three address variables and eight data variables).
After $2.0\times10^5$ trials, an average of $-620$ ($\approx70\%$ accuracy) accumulated reward is achieved by SUNA while NEAT reaches $-720$ ($\approx65\%$ accuracy).
However, SUNA continues improving its accuracy further and if more trials are allowed it would reach much higher accuracy.
NEAT, on the other hand, reaches rapidly its peak of accuracy without further improvements.
%This seems related to how the diversity of SUNA and NEAT work.
%While in NEAT good enough species are kept, SUNA preserves solutions that are different enough even when they are not sufficiently good compared with other solutions.
%In other words, NEAT focus on promising species that are somewhat different, while SUNA focus on sufficiently spread solutions and inside each niche of solutions only the best is kept.
%In SUNA, \textit{evolutionary pressure is strong within one niche but it is completely absent between niches}.
Ablation tests done in Section~\ref{ablation_tests} show that one of the features that allows SUNA to outperform NEAT is neuromodulation.
See Section~\ref{ablation_tests} for details.

\subsection{Function Approximation}

In this problem the algorithm has to learn a sequence given by the following equation:
\begin{equation}
	y = \frac{x^3}{1000} + 0.4x + 20\sin(\frac{x}{10}) + 20\sin(100x), 
\end{equation}
where $x$ and $y$ are respectively the input and output of the problem.
Moreover, the value of $x$ varies from $-100$ to $100$ in unit time steps.
The reward is the negative modulus of the difference between the agent's action (agent's estimation of $y$) and the correct sequence output $y$.

\begin{figure}[!ht]
\centering
 \includegraphics[width=0.5\textwidth]{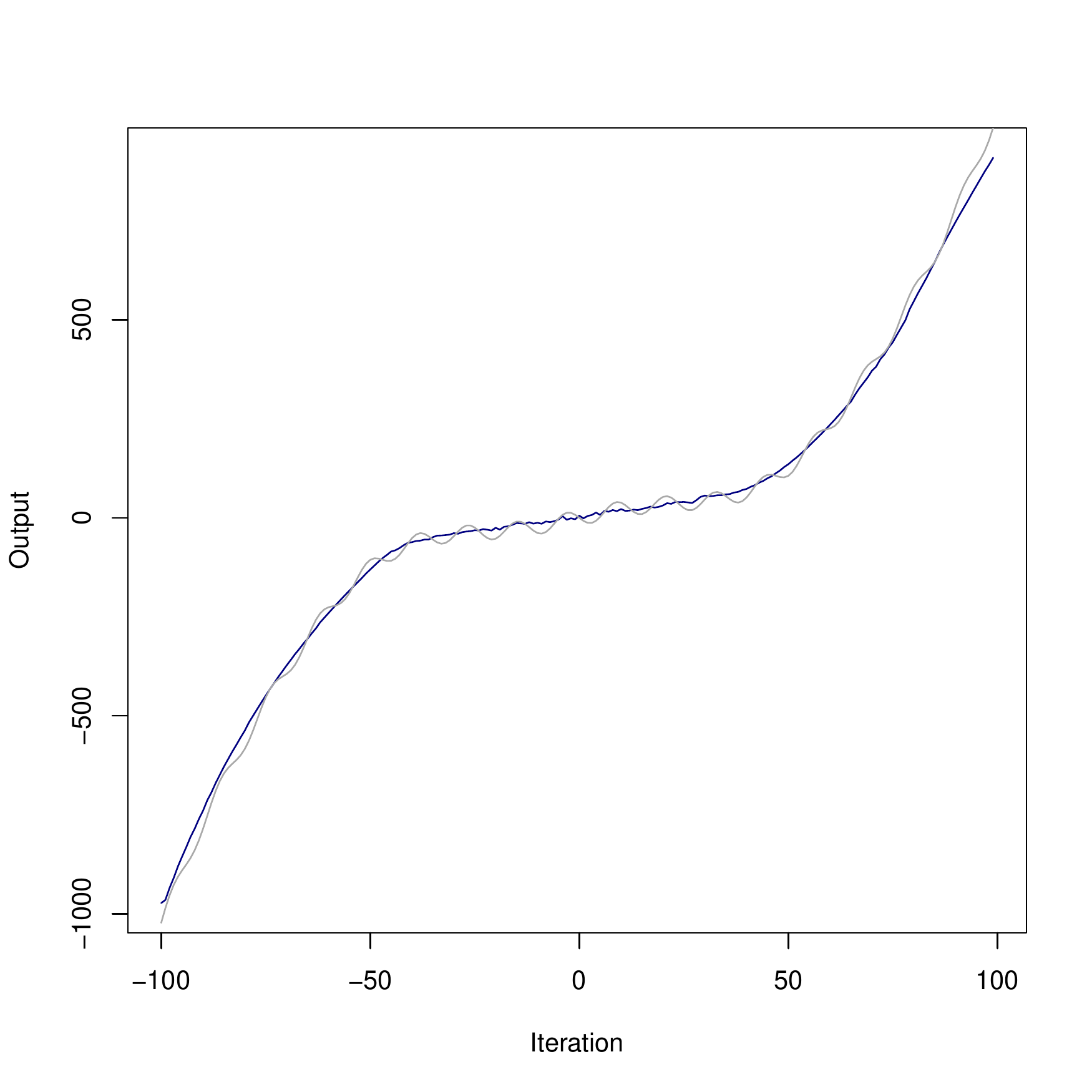}
 \caption{Average of the final learned sequence (dark blue) and the reference sequence (light gray) used to evaluate the fitness.}
 \label{fa_results}
\end{figure}

This problem cannot be solved properly by NEAT, since its input and output need to be real numbers of unknown range otherwise the problem is unreasonably easy.
Therefore, here only the SUNA's result is plotted (see Figure~\ref{fa_results}).
SUNA gets a very close approximation to the curve, Section~\ref{ablation_tests} will expose the importance of slow neurons and linear neurons in these classes of problems.

\subsection{Ablation Tests}
\label{ablation_tests}

%In contrast to the previous tests, {\bf here all the tests do not use any pre-processing system} such as the conversion of inputs/output to a certain range.
In contrast to the previous tests, {here all the tests do not use any pre-processing system} such as the conversion of input/output to a certain range.
So all input/output conversion necessary for using NEAT was excluded.
A natural consequence is that previous problems get more difficult.% with the results also changing.
To enable enough time for the algorithm to converge in these more difficult problems, runs have a maximum of $10^6$trials instead of $2\times10^5$ used in previous tests.
These types of tests were preferred to stress SUNA and verify its capabilities.

The ablation tests are composed of nine tests.
A control test and eight ablation ones.
Table~\ref{mean_final} shows the percentage in improvement or worsening after the ablation and Table~\ref{statistical_final} shows the statistical significance of the tests.
The following subsections will discuss every ablation in detail.

\begin{table*}[!ht]
\centering
\caption{Percentage of improvement (positive) or worsening (negative) of the ablation results in relation to the original one. The problems are Double Pole (DP), Function Approximation (FA), Mountain Car (MA), Multiplexer (MU) and Non Markov Double Pole (NMDP).}
\begin{tabular}{ |l|l|l|l|l|l|}
	%\hline
	%\multicolumn{3}{ |c| }{Team sheet} \\
	\hline
	% & GK & Paul Robinson \\ \hline
	Test Type & DP & FA & MC & MU & NMDP \\ \hline
	No control neuron & $0\%$ & $5.87\%$ & $0\%$ & $-10.55\%$ & $15.76\%$ \\
	No linear neuron & $0\%$ & $-22.38\%$ & $0\%$ & $-1.76\%$ & $11.18\%$ \\
	No neuromodulation & $0\%$ & $6.92\%$ & $0\%$ & $-66.17\%$ & $7.88\%$ \\
	No random neuron & $-6.28\%$ & $7.37\%$ & $0\%$ & $8.26\%$ & $-56.54\%$ \\
	No real weights & $-3.29\%$ & $9.43\%$ & $0\%$ & $6.16\%$ & $-99.85\%$ \\
	No sigmoid neuron & $0\%$ & $-3.73\%$ & $0\%$ &  $-3.35\%$ &  $1.97\%$ \\
	No slow neuron & $-6.46\%$ & $-32.69\%$ & $0\%$ & $12.61\%$ & $-35.94\%$ \\
	No threshold neuron & $-3.13\%$ & $6.43\%$ & $0\%$ & $1.56\%$ & $11.77\%$ \\
	\hline
\end{tabular}
\label{mean_final}
\end{table*}

\begin{table*}[!ht]
\centering
\caption{Statistical significance for the ablation tests. Two Mann-Whitney statistical tests were conducted for every ablation test in relation to the control one.
One statistical test verify if the results are better than the control one and the other statistical test verify the contrary. 
W or B means that the ablation has a final result which is respectively worse or better than the control one.
Only results significant enough (p-values below $0.05$) are considered. 
The problems are Double Pole (DP), Function Approximation (FA), Mountain Car (MA), Multiplexer (MU) and Non Markov Double Pole (NMDP).}
\begin{tabular}{ |l|l|l|l|l|l| }
	%\hline
	%\multicolumn{3}{ |c| }{Team sheet} \\
	\hline
	% & GK & Paul Robinson \\ \hline
	Test Type & DP & FA & MC & MU & NMDP \\ \hline
	No control neuron & $()$ & $()$ & $()$ & W$(0.01)$ & B$(0.04)$ \\
	No linear neuron & $()$ & W $(3.39e-6)$ & $()$ & $()$ & $()$ \\
	No neuromodulation & $()$ & B $(0.04)$ & $()$ & W $(2.16e-11)$ & $()$ \\
	No random neuron & $()$ & B $(0.03)$ & $()$ & B$(0.02)$ & W $(8.05e-5)$ \\
%	No random neuron & W$(0.08)$ & B $(0.03)$ & $()$ & B$(0.02)$ & W $(8.05e-5)$ \\
	No real weights & $()$ & B $(0.01)$ & $()$ & $()$ & W $(6.78e-12)$ \\
	No sigmoid neuron & $()$ & $()$ & $()$ &  $()$ &  $()$ \\
	No slow neuron & $()$ & W $(2.87e-6)$ & $()$ & B$(0.002)$ & W $(0.01)$ \\
%	No slow neuron & W$(0.08)$ & W $(2.87e-6)$ & $()$ & B$(0.002)$ & W $(0.01)$ \\
	No threshold neuron & $()$ & $()$ & $()$ & $()$ & $()$ \\
	\hline
\end{tabular}
\label{statistical_final}
\end{table*}

\subsubsection{No Linear Neurons}

Linear neurons are specially important in problems with real input/output and input/output with wide range.
The ablation test on function approximation confirms this affirmation, since the results degrade strongly when linear neurons are removed.
In fact, this result is statistically significant.

\subsubsection{No Control Neurons}

As explained before in Section~\ref{control_motivation}, with the introduction of control neurons it is now possible to create systems that switch on/off behaviors.
So the representation capabilities improved strongly but the tests with the ablation of control neurons show mixed results. 
Consequently, control neurons does not seem to be used to their maximum representation potential by the current evolutionary algorithm.% or they are just not needed at the current problems. 
%Therefore, in most of the problems not using them improves the performance, since this additional complexity only added to a more complicated search space and did not help the algorithm's representation 

There is a reason for the above.
Since certain groups of neurons work together forming functions, representation-wise, a higher level of control can be achieved if groups of neurons are activated or deactivated.
However, if the connections do not perfectly connect to all neurons involved in the function, instead of adding a higher level of control, the current level of control is disrupted and control neurons become instead another piece inside the monolithic block of neurons.
In other words, control neurons which are unwisely connected to the network, do not switch on/off behaviors (set of neurons) but just neurons. 

\subsubsection{No Neuromodulation}

The ablation tests show that neuromodulation is of utmost importance (a statistically significant $66\%$ worsening if removed) for the multiplexer problem. 
Actually, it is easy to see why neuromodulation is important in a problem like multiplexer that has input selection as its main task, because neuromodulation does exactly that, it modulates a node based on another one. 

\subsubsection{No Random Neurons}

Random neurons may seem at first glance somewhat weird, summing with the fact that they are of extreme importance for the non-Markov double pole problem is yet more perplexing.
However, notice that random neurons are exactly important on problems that require some form of exploration of the environment (e.g., non-Markov double pole) while being irrelevant where no exploration is necessary (function approximation and multiplexer).

It is surprising that even without any kind of exploration dynamics hard-coded, that the algorithm develop by itself, at least initially, dynamics that explore the environment.

\subsubsection{No Real Weights}

This time, tests verify the importance of using real weights, i.e., these ablation tests use only -1, 1, or neuromodulation (another neuron's output) as possible weight values. 
The results show that setting the non-Markov double pole balancing aside, in all problems the ablated algorithm's performance is similar to when real weights are used, with the double pole balancing problem having improved a bit the results after the removal of real weights. 
However, for the non-Markov double pole balancing problem the removal of real weights causes substantial decrease in performance.

This can be explained by the following reasoning. 
The absence of real weights is a constraint  added to the problem and when this constraint helps to find a good solution (in the case of double pole balancing) it is surely welcome, but sometimes it makes impossible to find good solutions (non-Markov double pole balancing). 
Notice that the only difference between these two problems (non-Markov double pole and double pole) is the presence/absence of the three velocity variables and therefore when they are not present it is necessary to construct them with recurrence and some calculation, the recurrence is present but it seems that this calculation is never precise enough without real weights and therefore the algorithm is never able to solve the non-Markov version.  

\subsubsection{No Sigmoid Neurons}

The removal of sigmoid neurons is shown to have no impact on the system performance. 
All the comparisons had a small percentage difference as well as were not statistically significant. 
This may be related with the similarity between sigmoid and threshold neurons. 
Sigmoid neurons have a non modifiable steep curve given by a hyperbolic tangent which has the same output for almost all inputs, aside from small portions of the input range near zero.
Therefore, the system does not suffer from losing one or the other.

\subsubsection{No Slow Neurons}

Slow neurons are important when the information needs to be preserved from previous iterations. 
That is why they are very important to non-Markov double pole and function approximation problems. 
Notice that slow neurons allow for the computation of an average of its input (i.e., the input gets accumulated and is naturally averaged in its output), therefore they are good estimates of rates of variance as well as derivatives.

\subsubsection{No Threshold Neurons}

As mentioned before, threshold neurons and sigmoid neurons have similar activation functions. 
So removal of one or the other does not impact on the performance.
This justifies not statistically significant results as well as small percentage difference.

\section{The Laws of Learning}

It was shown that each problem class requires some features from learning algorithm in order to be tackled.
However, these features are not specific to each problem class.
Many problem classes require similar learning features.
Therefore, the question of whether there is a minimal set of features required to solve all problems is raised.
%\footnote{The authors are familiar with the No Free Lunch Theorem (NFLT), however notice that the conditions for the NFLT to hold does not apply to algorithms that uses information of the problem to change itself \cite{}}.

Thus, similarly to the laws of Physics, which are the minimum set of dynamics that describe the universe.
Are there a minimum set of learning features (neuron types, recurrency, etc...) that would allow a learning algorithm to learn any problem?

We think there is a minimum set. 
In this work we identified some important learning features.
But there are certainly many more that need to be identified and unified.

\section{Extensions to NEAT and Extensions to SUNA}

NEAT has various extensions, such as FS-NEAT \cite{ethembabaoglu2008automatic} where the initial topology is set to only one input connecting to the output instead of the initial fully connected topology between inputs and outputs, RBF-NEAT \cite{kohl2008evolving} which uses an additional mutation that adds Radial Basis Function (RBF) nodes, Cascade-NEAT \cite{kohl2012integrated} where only the connections to the last added node are evolved and SNAP-NEAT \cite{kohl2012integrated} which integrates mutations of three algorithms (NEAT, RBF-NEAT and Cascade-NEAT) into one.

Regarding indirect encoding, Compositional Pattern Producing Networks (CPPN) is an abstraction of natural development that uses neural networks to create an indirect encoding instead of developmental encodings.
Usually, CPPNs are developed with the algorithm called CPPN-NEAT that uses NEAT to evolve it \cite{stanley2007compositional}.
Moreover, HyperNEAT is an indirect encoding based algorithm which uses CPPN-NEAT to create patterns of weights in substrates \cite{stanley2009hypercube}.
%However, these substrates are manually defined and current advances tries to tackle this problem with automatic development of substrates.
%\cite{huizinga2014evolving}.

In fact, many of the NEAT extensions can be easily applied to SUNA.
%This happens because SUNA has a different way of evolving but still shares the main neuroevolutionary paradigm.
For example:
\begin{itemize}
	\item Similarly to RBF-NEAT, RBF nodes can be included in SUNA;
	\item Connections to old nodes can be frozen, having only the connections to recent added nodes evolve (similar to Cascade-NEAT);
	\item Additional mutation operators can be used or mixed, integrating other approaches (SNAP-NEAT);
	\item CPPNs can be developed using SUNA in the same way that they were developed with NEAT, using a CPPN-SUNA algorithm;
	\item HyperSUNA can be developed using either a CPPN-SUNA or by substituting CPPN with SUNA entirely.
\end{itemize}

\section{Conclusions}

This paper proposed a new TWEANN that joins most of the neural network features into one unified representation.
In fact, a natural consequence of the unified representation is a much more powerful representation power.
This affirmation was verified by the tests where the proposed algorithm outperformed NEAT on most of the problems.%, having similar results on the remaining ones. 
Moreover, ablation tests demonstrated that good results can only be obtained with the presence of the added new features.
%Each problem requires some specific features in the learning algorithm to be learned.
Therefore, this work give steps forward an algorithm that can learn any problem and we raise the question of if there is a minimal set of features that can let an algorithm learn anything.

Actually, representation alone is not enough, a diversity preserving method that called Spectrum Diversity is also necessary.
%the proposed Spectrum Diversity method achieved a good balance between diversity and selection pressure while scaling as much as the designer wants it to scale with the size of the chromosome.
Spectrum Diversity allows novel enough chromosomes to be kept even when their fitness is abysmally poor. 
Consequently, this allows different approaches to develop without the deleterious competition of faster evolving ones.
Moreover, Spectrum diversity scales better with the size of the chromosome than Speciation because of its use of chromosome spectra instead of the genes themselves.
This diversity method together with the chromosome spectrum concept should find uses in many other algorithms, specially the ones that utilize a great variety of gene types in their genotype as well as applications where the chromosomes are high dimensional.

Thus, the current work sheds light on a new representation as well as a new diversity keeping method and its associated spectrum concept, opening up arguably better possibilities for the creation of algorithms.
Moreover, the overall good results motivates further applications of the approach to real-world problems and extensions such as an indirect encoding version.
%looks promising.

%\begin{itemize}
%\end{itemize}

%\section*{Acknowledgment}


\ifCLASSOPTIONcaptionsoff
  \newpage
\fi

\bibliographystyle{IEEEtran}
\bibliography{../sigproc,deep_neural_networks}


%\vfill


\end{document}